\documentclass[11pt]{article}
\usepackage[letterpaper,margin=1in]{geometry}

\usepackage{amsmath} 
\usepackage{amssymb}  
\usepackage{graphicx}
\usepackage{epstopdf}
\usepackage{xcolor}
\usepackage{microtype}
\usepackage{tabularx}
\usepackage{float}
\usepackage{comment}

\usepackage[colorlinks,citecolor=black,urlcolor=black,bookmarks=false,hypertexnames=true, allbordercolors=black, allcolors=black]{hyperref}

\newcommand{ \vctwo }[2] {
 \left( \begin{array}{c }
  #1        \\
  #2
\end{array} \right)  }

\newcommand{\txt}[1]{ \hspace{1em}\mbox{#1}\hspace{1em} }

\title{ \Large \bf
Minimum Time Trajectories for a Car-Like Mobile Robot \\[4pt]
Moving with Rigid Wheels Under Non-Sliding Constraints }
\author{
	J. Z. Ben-Asher$^{1}$, E. D. Rimon$^{2}$ and  L. A. Ravina$^{2}$
	\thanks{$^{1}$Dept. of AE, Technion, Israel. 
 $^{2}$Dept. of ME, Technion, Israel. 
	}   
}

\usepackage[colorlinks,citecolor=black,urlcolor=black,bookmarks=false,hypertexnames=true, allbordercolors=black, allcolors=black]{hyperref}
\usepackage{cite}

\def \eex{ \hfill $\circ$ }
\date{Technical Report \\ \today}
\begin{document}
\maketitle
\pagestyle{plain}
\sloppy

 \noindent {\large\bf Abstract} {\it This paper studies the minimum time~trajectories~of a car-like mobile robot navigating in an~obstacle~free~environ\-ment. The robot, with forward and backward speeds, is controlled by bounded front-wheels acceleration and limited front-wheels steering rate. The paper extends previous results which solved this problem for the kinematic car-like robot. However the kinematic model assumes pure rolling~at~the~wheels~ground cont\-acts. This assumption requires
non-sliding constraints for the front and rear wheels that
can only be handled by the robot dynamics. This paper formulates
the non-sliding constraints based on the robot dynamics
then augments the kinematic model time-optimal path primitives with three new path primitives associated with the non-sliding constraints. The three non-sliding path primitives together with the kinematic model twelve path primitives
form the
car-like robot time optimal trajectories.
Approximate analytic~\mbox{solutions}~for~the~non-sliding path primitives  are also provided.
Examples study the time-optimal path primitives along representative maneuvers, illustrating how the non-sliding constraints influence the time optimal trajectories of the car-like robot. }

\section{Introduction}

\noindent This paper studies the time optimal trajectories~of~a~\mbox{car-like} mobile robot~nav\-igating in an~obstacle free planar environment. Time optimal trajectories for mobile robots  have been studied for several decades~\cite{modern_robotics}.  Dubin's classical work, published in 1957~\cite{dubins}, was the first to study the time-optimal paths for a~constant speed car moving in an obstacle-free environment and to formulate the exact optimal solution.  Reeds and  Shepp~\cite{reeds&shepp} extended Dubins' work to a~car-like robot that moves with constant speed 
forwards and backwards. For a~variable speed car-like robot,  this problem~was~form\-ulated in a so-called {\em simplified  kinematic model}~\cite{modern_robotics}. The car-like mobile robot is simplified into a~unicycle having no steering front wheels with speed and turn-rate  as control inputs. Using the control inputs in a~manner 
that limits 
path curvature, several decades of research studied the time optimal trajectories under the  simplified  kinematic model. See Boissonnat~\cite{boissonnat92,boissonnat94},  Sussman~\cite{sussmann&tang}, Sou\`{e}res~\cite{soueres&laumond,soueres&boissonnat},
Fraichard~\cite{fraichard&cheuer}, Wang~\cite{wang2009}, {Wolek~\cite{cliff_2016} and Zhang \cite{zhang_car}.



A  realistic car-like 
robot model having body-fixed rear wheels and steering front wheels was suggested by Laumound~\cite{laumond94}. The control inputs in this model are bounded front-wheel linear acceleration and limited front-wheels steering rate. However,  being concerned with obstacle avoidance, 
the model in~\cite{laumond94} was immediately substituted by the simplified  kinematic model.
Ref.~\cite{yossi_car}, which the current paper extends, employed Pontryagin's {\em minimum principle} and singular control theory to obtain 
the time optimal path primitives for a~car-like  robot navigating in an~obstacle free environment, using the full kinematic model of  
Ref.~\cite{laumond94}. While analyzing the problem, Ref.~\cite{yossi_car} obtained {\em twelve path primitives} that form 
the  time optimal paths
assuming that the robot wheels are not sliding.
Some primitives are generated by {\em regular controls,}
determined directly from the minimum principle. Other primitives are governed by {\em singular controls,} determined by the vanishing of a switching function and its time derivatives. 


The optimal control techniques used in 
this paper have been applied to other mobile robot motion planning problems. Examples include car-like robot 
path tracking~\cite{xie_2021}, driverless car  
maneuvers in congested highway traffic~\cite{zhou_2019} and Dubin's car trajectories among moving obstacles~\cite{shima_2022}. Simultaneous efforts extended sampling based methods to 
plan the motion of wheeled mobile robots in the presence of obstacles~\cite{orthey2023sampling,palmieri2017,palmieri2016}. The time optimal path primitives described in the current paper can be incorporated into sampling based planners for a~more efficient search of the mobile robot states.
Heiden~\cite{heiden_ral21} 
developed a~benchmark package
that evaluates car-like robot paths in the presence of obstacles based on path length, curvature 
and clearance from obstacles. The time optimal path primitives described in the current paper can be incorporated into {\small Bench-MR} for a more complete offering of car-like robot motion planning methods in the presence of obstacles.

{\bf Paper contributions:} This paper adds dynamics based non-sliding constraints to the kinematic~car-like robot model in order to characterize the time optimal path primitives associated with the 
non-sliding 
constraints. To this end, the car-like robot rigid-body dyn\-amics is formulated under simplifying assumptions~\mbox{leading}~to the {\em bicycle model}~\cite{paden&frazzoli2016}.  The non-sliding constraints form quadratic 
inequalities in the 
robot  control inputs. When comb\-in\-ed with the need to maintain positive normal~loads~at~the wheels ground contacts, the allowed control set forms a~{\em compact convex set.} This 
means that time opt\-imal paths always exist (usually appearing as several symmetrically arranged paths), and the optimal control action can be uniquely determined at each time instant along the robot path. The paper analyzes the time optimal control problem under this 
convex 
control set and obtains the time optimal path primitives associated with the non-sliding constraints. The paper then verifies the 
time optimal primitives using direct numerical solvers,
showing with examples how the non-sliding constraints influence the car-like robot time optimal trajectories along representative maneuvers.

The time optimal path primitives described in this paper can be used to improve existing mobile robot motion planners. One important way would be to incorporate the path primitives into kinodynamic planners~\cite{panasoff_RSS25}. These planners construct a~tree in the mobile robot state space (position and velocity) whose edges represent dynamically feasible mobile robot paths taken under sampled control actions. The selection of control action 
can be informed by the path primitives and their 
solutions described in this paper.
Equally promising would be to incorporate the path primitives into machine learning planners. One can generate data sets of time optimal maneuvers using direct numerical solvers. These maneuvers can be automatically parsed into their time~optimal path primitives, 
then used to train a~{\em path primitive classifier} that predicts the next path primitive and its switch time based on the mobile robot current state and current~path primitive. 


The paper is organized as follows. Section~II~describes modeling assumptions and the car-like robot dynamic model. Section~III introduces the non-sliding constraints and formulates the
time optimal control problem.  Section~IV shows~that the non-sliding constraints form a~convex set of allowed control inputs.
Section~V  provides the main analysis of the time optimal control problem. Section~VI describes representative numerical examples that verify the analysis. The conclusion 
suggests future research topics that use the time optimal path primitives. An appendix describes approximate analytic solutions for the non-sliding path primitives.

\section{Modeling and Problem Formulation}


\noindent This paper considers a front wheel drive car-like 
robot navigating on a~flat horizontal floor (Fig.~\ref{fig:CarStates}). The robot
state variables are the
rear wheels midpoint position, {\small $(X,Y)$}, 
the robot heading angle, $\theta$, the front wheels steering angle, $\phi$, and the front wheels midpoint linear speed $\nu$ (Fig.~\ref{fig:CarStates}). Thus
\[
\mbox{\small $ 
\mbox{\small $S$}(t) = \big( \mbox{\small $X$}(t), \mbox{\small $Y$}(t),  \mbox{\small $\theta$}(t),  \phi(t), \nu(t) \big)
$} 
\]
\noindent where the front-wheels state variables are bounded by $|\phi(t)| \leq \phi_{max}$ and $|\nu(t)|\leq \nu_{max}$, with $\phi_{max} < \pi/2$.
{Note that the car-like robot is both forward and backward moving.}
The control inputs are the front wheels linear acceleration
and the front wheels steering rate.
Thus
\[ 
\mbox{\small $
   u_1(t) \!= \! \dot{\nu}(t) \hspace{1.em} \mbox{and} \hspace{1.em} u_2(t) \!=\! \dot{\phi}(t) $} 
\]
\noindent where the controls 
are independently bounded by the limits
\begin{equation}\label{eq:maximalCrossTrackAcceleration}
\mbox{\small $
|u_1(t)| \leq a_{max}  \hspace{.5em}  \mbox{and} \hspace{.5em} |u_2(t)| \leq b_{max}. $} 
\end{equation}

 \begin{figure}
 	\centering
 	\includegraphics[width=0.65\linewidth]{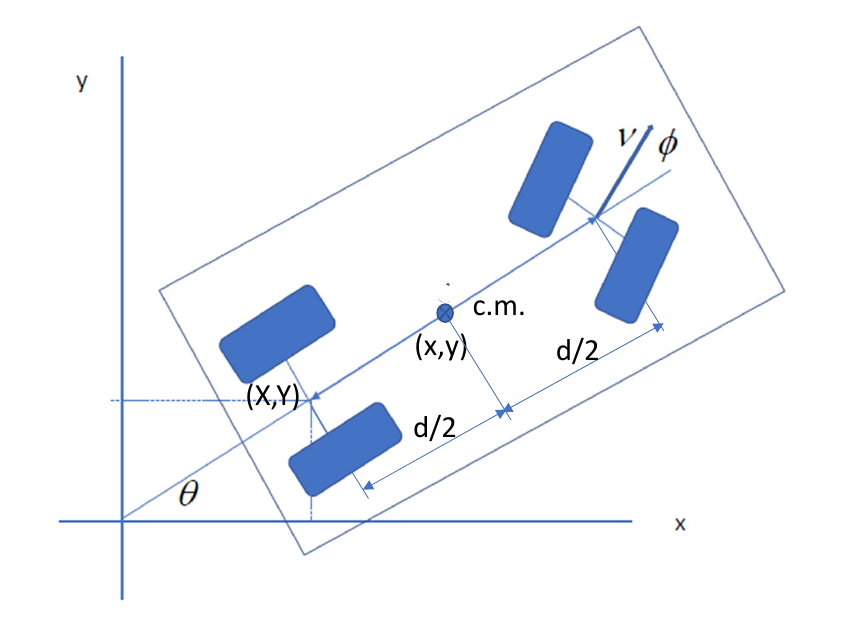}
\caption{The car-like mobile robot state variables are $(\mbox{\scriptsize  $X$},\mbox{\scriptsize  $Y$},\mbox{\footnotesize  $\theta$})$, front wheels steering angle $\phi$ and front wheels midpoint speed $\nu$.} \label{fig:CarStates}
\end{figure}

\noindent Based on the car-like robot kinematics (Fig. \ref{fig:CarStates}), the top-view position of the
rear wheels midpoint {\small $(X,Y)$} and the robot heading angle $\theta$ satisfy the kinematic equations~\cite{laumond94,yossi_car}
\begin{equation}\label{eq:RobotEOM1}
\begin{split}
\mbox{\small $\dot{X}$}(t) &= \nu(t)\cos \phi (t)\cos \theta (t)\\
\mbox{\small $\dot{Y}$}(t) &= \nu(t)\cos \phi (t)\sin \theta(t)\\
d \!\cdot\!  \dot{\theta} (t) &= \nu(t)\sin\phi (t)
\end{split}  
\end{equation}
\noindent where $d$ is the distance between the rear and front wheels midpoints.
(Fig.~\ref{fig:CarStates}). 
Note that {\small $\dot{\theta}$} measures rotation rate  about any body point and in particular about the car-like robot center of mass. 


The kinematic 
equations are next used to formulate the car-like robot dynamic equations.  The top-view position~of~the~car-like robot
center of mass in the world frame, $(x,y)$,~is~given~by
\[ 
\begin{pmatrix}
x(t) \\ y(t)
\end{pmatrix} = 
\begin{pmatrix}
\mbox{\small $X$}(t) \\ \mbox{\small $Y$}(t)
\end{pmatrix} +
\frac{d}{2}
\! \begin{pmatrix}
\cos \theta(t) \\ \sin \theta(t)
\end{pmatrix} \, .
\]
\noindent The car-like robot center of mass velocity  is thus
\begin{equation} \label{eq.com_vel}
\mbox{\small $  \begin{pmatrix}
\dot{x} \\ \dot{y}
\end{pmatrix} =\nu \cos \phi 
\!\cdot\! \!
\begin{pmatrix}
\cos \theta \\ \sin\theta
\end{pmatrix} +
\mbox{\normalsize $\frac{d}{2}$}  \dot{\mbox{\small $\theta$}}   \!
\begin{pmatrix}
-\sin\theta \\ \cos\theta
\end{pmatrix} $} 
\end{equation}
\noindent and the center of mass acceleration is
\begin{equation} \label{eq.com_acc}
\mbox{\small $
\begin{split}
\begin{pmatrix}
\ddot{x} \\ \ddot{y}
\end{pmatrix} = &  \big( \dot{\nu} \cos \phi - \nu \dot{\phi} \sin \phi - \frac{d}{2}   \dot{\mbox{\small $\theta$}}^2 \big) \!\cdot \! \!
\begin{pmatrix}
 \cos \theta \\ \sin\theta
\end{pmatrix} \\
& +  (\nu \dot{\mbox{\small $\theta$}} \cos \phi + \frac{d}{2}  \ddot{\mbox{\small $\theta$}}) \!\cdot\! \!
\begin{pmatrix}
-\sin \theta \\ \cos \theta
\end{pmatrix} .
\end{split} $} 
\end{equation}
\noindent Substituting the controls $u_1 \!= \! \dot{\nu}$ and $u_2 \!=\! \dot{\phi}$ and the kinematic relation $d \!\cdot\! \dot{\mbox{\small $\theta$}} \!=\! \nu \sin\phi $ gives the center of mass acceleration expressed in terms of the state variables and controls:
\begin{equation} \label{eq.acc_world}
\mbox{\small $
\begin{split}
\begin{pmatrix}
\ddot{x}  \\ \ddot{y}
\end{pmatrix} =&  \big( u_1 \cos \phi - u_2 \nu  \sin \phi - \frac{ \nu^2}{2d} \sin^2\phi \big) \!\cdot\! 
\begin{pmatrix}
 \cos \theta \\ \sin\theta
\end{pmatrix} \\
& +  \big(  \frac{\nu^2}{2d}  \sin(2\phi) + \frac{d}{2} \ddot{\mbox{\small $\theta$}} \big) \!\cdot\! 
\begin{pmatrix}
-\sin \theta \\ \cos \theta
\end{pmatrix}
\end{split} $} 
\end{equation}
\noindent where $d \!\cdot\! \ddot{\theta}$ is given by
\[
\mbox{\small $
 d \!\cdot\! \ddot{\theta} 
=  \dot{\nu} \sin \phi + \nu \dot{\phi} \cos\phi = u_1 \sin \phi + u_2  \nu \cos\phi$} \, .
\]

\begin{figure} \centering
\includegraphics[width=0.65\linewidth]{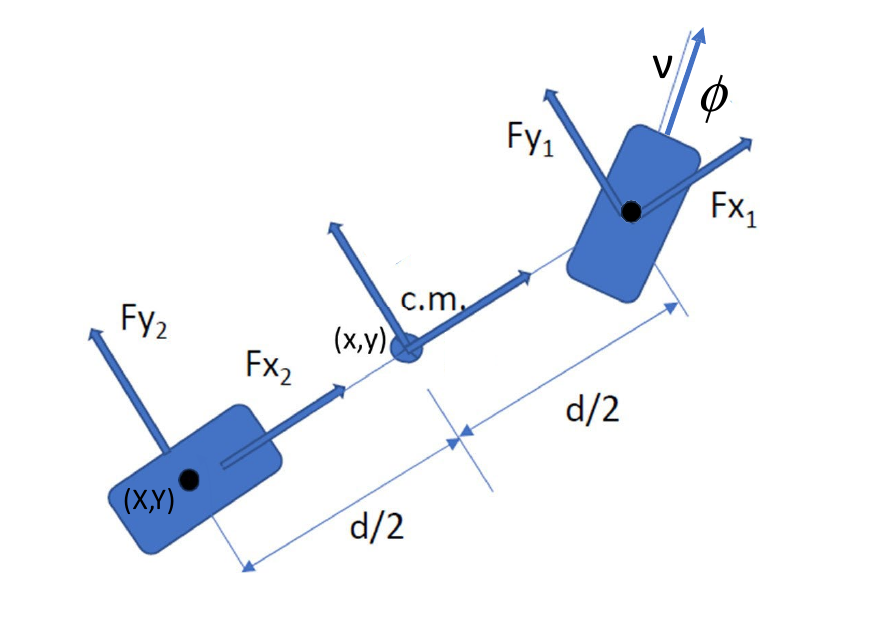}
\caption{The simpler bicycle-like mobile robot has the same state variables: $(\mbox{\scriptsize  $X$},\mbox{\scriptsize  $Y$},\mbox{\footnotesize $\theta$})$, front wheel steering angle $\phi$ and front wheel speed $\nu$.} \label{fig:model}
\end{figure}

The forces and moments that act on the car-like robot are next written in body coordinates.  Consider the car-like robot {\em bicycle
model} depicted in Fig.~\ref{fig:model}. The front wheels are lumped into a~steering wheel with ground forces
$(F_{x_1},F_{y_1})$.  The rear wheels are lumped into a~body fixed wheel with ground forces $(F_{x_2},F_{y_2})$.
Based on Eq.~\eqref{eq.acc_world}, the bicycle model  dynamic equations in body coordinates are given by
\begin{equation} \label{eq:Bicycles}
\mbox{\small $ 
\begin{split}
& m \!\cdot\! \big( u_1 \cos \phi \!-\!  u_2 \nu \sin \phi \!-\!  \tfrac{\nu^2}{2d}  \sin^2 \phi \big)  = F_{x_1} \!+\! F_{x_2} \\
 & m \!\cdot\!  \big( \tfrac{\nu^2}{2d} \sin (2\phi) + \tfrac{1}{2} ( u_1 \sin \phi + u_2 \nu \cos \phi  ) \big) = F_{y_1} \!+\! F_{y_2} \\
&  \mathcal{I}  \!\cdot\! \ddot{\theta}         
= \tfrac{1}{d} \mathcal{I} \!\cdot\! ( u_1 \sin \phi + u_2 \nu \cos \phi ) = \tfrac{d}{2} \!\cdot\! ( F_{y_1} \!-\!  F_{y_2} )
\end{split} $} 
\end{equation}
\noindent where $m$ is the robot mass and $\mathcal{I}$ is the robot moment~of inertia measured about 
its~center of mass.\\
\indent {\bf Bicycle model insight:} Consider the bicycle model dynamics described in Eq.~\eqref{eq:Bicycles}. Using Eq.~\eqref{eq.com_vel}, the bicycle's center of mass linear velocity in body coordinates is given by $\mathrm{v}_{cm} \!=\! (\cos\phi,\tfrac{1}{2} \sin\phi)$. Using
$\mathrm{v}_{cm}$ and $\mathrm{v}^{\perp}_{cm} \!=\! (-\tfrac{1}{2} \sin\phi,\cos\phi)$,
the bicycle model  dynamic equations can be written in body coordinates as
\begin{equation} \label{eq:Bicycles1}
 \mbox{\small $ 
 m \!\cdot\! \left( u_1  \mathrm{v}_{cm} + u_2 \nu \dot{\mathrm{v}}_{cm} + 
 \tfrac{\nu^2}{2 d} \mathrm{v}^{\perp}_{cm} \right) = \vctwo{\!\! F_{x_1} \!+\! F_{x_2} \!\! }{\!\!  F_{y_1} \!+\! F_{y_2} \!\! }
 $} 
 \end{equation}
\noindent  and
\begin{equation} \label{eq:Bicycles2}
  \mbox{\small $
 \tfrac{1}{d} \mathcal{I} \!\cdot\! ( u_1 \sin \phi + u_2 \nu \cos \phi ) = \tfrac{d}{2} \!\cdot\! ( F_{y_1} \!-\!  F_{y_2} ) $} 
\end{equation}
\noindent The term proportional~to $\nu^2$ 
represents centrifugal forces~while 
all other terms 
depend linearly on the controls and represent linear-and-angular momentum 
effects.~\eex\\
\indent Eq.~\eqref{eq:Bicycles} specifies three constraints on the ground forces.
The rear wheel of the bicycle model is unactuated. When this wheel is rolling without slippage, 
it satisfies the kinematic constraint $r_w \!\cdot\! \dot{\psi} \!=\! \nu \cos \phi$, where
$r_w$ is the wheel radius and $\psi$ is the wheel rotation angle. The rear wheel axial torque is $r_w \!\cdot\! F_{x_2}$ and its axial dynamics is given by
\begin{equation} \label{eq.rear0}  
\mbox{\small $
 \mathcal{I}_w \!\cdot\! \ddot{\psi} = \frac{\mathcal{I}_w}{r_w} \!\cdot\!  ( u_1\cos \phi - u_2 \nu \sin \phi )  = - r_w \cdot F_{x_2} 
$} 
\end{equation}
\noindent where $\mathcal{I}_w$ is the 
rear-wheel axial moment of inertia. 
This  moment of inertia can be written as $\mathcal{I}_w \!=\! m_w \rho^2_w$, where $m_w \!<<\! m$ is the rear wheel mass and $\rho_w$ the wheel radius of gyration. For simplicity assume $\rho_w \!\cong\! r_w$.
Substituting $\mathcal{I}_w  \!=\! m_w r^2_w$  in Eq.~\eqref{eq.rear0} gives
\begin{equation} \label{eq.rear}
\mbox{\small $ 
F_{x_2} =  - m_w  \cdot ( u_1 \cos \phi - u_2 \nu \sin \phi ).
$} 
\end{equation}
\noindent Eqs.~\eqref{eq:Bicycles} and~\eqref{eq.rear} specify~the~ground~forces~\mbox{\small $(F_{x_1},F_{y_1})$}~and \mbox{\small $(F_{x_2},F_{y_2})$} in terms of the state variables and controls:
\begin{equation}\label{eq:forces}
\mbox{\small $
\begin{split}
 F_{x_1} &= (m \!+\! m_w) \!\cdot\! ( u_1 \cos \phi - u_2 \nu \sin \phi)  - m \!\cdot\! \frac{\nu^2}{2d} \sin^2 \phi \\[-2pt]
 F_{y_1} &= \tfrac{m}{4} \!\cdot\! \left( \tfrac{\nu^2}{d} \sin(2\phi) + ( \mbox{\small $1$} \!+\! \mbox{\small $4$} \tfrac{\rho^2}{d^2} ) ( u_1 \sin \phi + u_2 \nu \cos \phi  ) \right)  \\[-2pt]
 F_{x_2} &= - m_w  \cdot ( u_1 \cos \phi - u_2 \nu \sin \phi )\\[-2pt]
  F_{y_2} &= \tfrac{m}{4} \!\cdot\! \left( \tfrac{\nu^2}{d} \sin(2\phi) + ( \mbox{\small $1$} \!-\! \mbox{\small $4$} \tfrac{\rho^2}{d^2} ) ( u_1 \sin \phi + u_2 \nu \cos \phi  )\right) 
\end{split} $} 
\end{equation}
\noindent where we substituted $\mathcal{I} \!=\! m \rho^2$, where $m$ is the robot mass and  $\rho$ is the robot
radius of gyration measured about a~vertical axis passing through its center of mass.

{\bf Caveat with bicycle model:}   {The bicycle model ignores how the normal loads split between the underlying car-like robot left and right wheels.
This paper assumes that the longitudal load transfer dominates the lateral load transfer, due to the fact
that the left-right wheel base is significantly shorter than the front-rear wheel base in typical car-like robots (Fig.~\ref{fig:CarStates}).~\eex}

\section{Non-sliding Constraints}

\noindent This paper assumes a car-like robot that moves with \emph{rigid wheels} on a~flat horizontal floor. The non-sliding constraints are now formulated for the 
robot's bicycle~model. Each 
wheel of the bicycle model is assumed to maintain a~frictional point contact that satisfies the Coulomb friction law.
The Coulomb friction law consists of tangential ground forces and moment about the wheels contact normals. For simplicity assume
negligible torsional 
friction and hence negligible moment  about the wheels contact normals.
The tangential coefficients of friction,
$\mu_x$ and $\mu_y$,  form an~elliptic friction cone whose
axes are aligned with the respective wheel of the bicycle model. 
The {\em rear wheel} non-sliding constraint takes the elliptic form
\begin{equation} \label{eq.nonslide}
\mbox{\small $ 
\frac{1}{\mu^2_x} F^2_{x_2} + \frac{1}{\mu^2_y} F^2_{y_2} \leq (N_2)^2
$} 
\end{equation}
\noindent where $N_2$ is the {\em normal load} at  the rear wheel ground contact (Fig.~\ref{fig:side.cropped}).  The {\em front wheel} direction is $(\cos\phi,\sin\phi)$ in body coordinates, and its non-sliding constraint takes the form
\begin{equation} \label{eq.nonslide1}
\mbox{\small $ 
  \frac{1}{\mu^2_x} (F_{x_1} \cos\phi  \!+\!  F_{y_1} \sin\phi)^2  \!+\!
\frac{1}{\mu^2_y} (F_{y_1} \cos\phi  \!-\!  F_{x_1} \sin\phi ) ^2 
\leq (N_1)^2  
$} 
\end{equation}
\noindent where $N_1$ is the {\em normal load} at  the front wheel ground contact (Fig.~\ref{fig:side.cropped}).

\begin{figure}
	\centering
\includegraphics[width=0.65\linewidth]{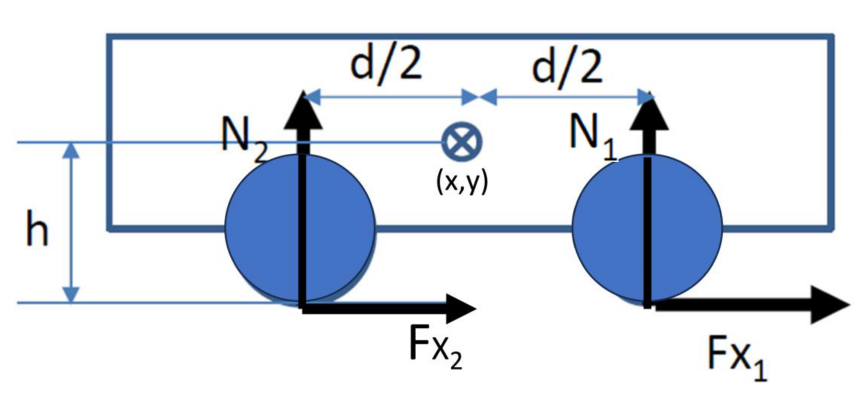}
\caption{Side view of the bicycle-like mobile robot showing the normal loads $N_1$ and $N_2$ at the ground contacts.}
	\label{fig:side.cropped}
\end{figure}

{\bf Normal load constraints:} The normal loads $N_1$ and $N_2$ balance the robot mass against gravity
while maintaining equilibrium of moments about a~horizontal y-axis that passes through the robot center of mass (Fig. \ref{fig:side.cropped}). Thus
\[
\mbox{\small $
N_1 + N_2 = m g  \txt{and} \frac{d}{2} N_1 + (F_{x_1} \!+\! F_{x_2})  h - \frac{d}{2} N_2 =0 $} 
\]
\noindent where $h$ is the height of the robot center of mass above ground. The two equations give the normal loads:
\begin{equation} \label{eq.nlodas}
\mbox{\small $\!\!\!
N_1 \!=\! \tfrac{1}{2} mg - \frac{h}{d} (F_{x_1} \!+\! F_{x_2}) 
\,\, \mbox{and} \,\,
N_2 \!=\! \tfrac{1}{2} mg + \frac{h}{d} (F_{x_1} \!+\! F_{x_2}). $} 
\end{equation}
\noindent The front-wheel normal load thus \emph{decreases} when the car-like robot is accelerating and \emph{increases} when the car-like robot is decelerating. The rear-wheel normal load \emph{increases} when the car-like robot is accelerating  and \emph{decreases} when the car-like robot is decelerating. The car-like robot must maintain both wheels of its bicycle model engaged against the ground. This gives two additional inequality constraints 
\begin{equation} \label{eq:normal}
\mbox{\small $
\tfrac{1}{2} mg \!-\! \frac{h}{d} (F_{x_1} \!+\! F_{x_2}) \!\geq\! 0 
\,\,\,\mbox{and}\,\,\, \tfrac{1}{2} mg \!+\! \frac{h}{d} (F_{x_1} \!+\! F_{x_2}) \!\geq\! 0 $} 
\end{equation}
\noindent where \mbox{\small $(F_{x_1},F_{y_1})$} and \mbox{\small $(F_{x_2},F_{y_2})$} are functions of the state variables and controls according to Eq.~\eqref{eq:forces}. The {\em full non-sliding constraints} thus include 
the quadratic constraints 
\eqref{eq.nonslide}--\eqref{eq.nonslide1}~and~the~\mbox{linear~}\mbox{constraints}~\eqref{eq:normal}
expressed in terms of the state and control variables.


{\bf The time-optimal control problem:}
Consider two endpoints of states $S(0)$ and $S(t_f)$.
Find the state-space path $S(t) $ for $t \epsilon [0,t_f]$ connecting these endpoints,~and~the corresponding control inputs $(u_1(t),u_2(t))$ for $t \in\!\! [0,t_f]$~that~minimize~the total travel time, ${t_f}$, subject to the control limits \eqref{eq:maximalCrossTrackAcceleration}, the non-sliding constraints~\eqref{eq.nonslide}--\eqref{eq.nonslide1},
the normal load constraints~\eqref{eq:normal} and the state variables constraints.

\section{Convexity of Non-sliding Constraints} 

\noindent This section shows that the full non-sliding constraints 
form a convex set of allowed control inputs.
Assume for simplicity a~uniform coefficient of friction,
$\mu \!=\! \mu_x \!=\! \mu_y$. Substituting for $N_1$ and~$N_2$ according to Eq.~\eqref{eq.nlodas},
the non-sliding constraints 
\eqref{eq.nonslide}-\eqref{eq.nonslide1}
take the simpler form
\begin{equation} \label{eq.fw_quad}
\mbox{\small $
F_{x_1}^2 \!+\! F_{y_1}^2  \leq  \mu^2 \!\cdot\!  \big( \tfrac{1}{2} mg - \tfrac{h}{d} (F_{x_1} \!+\! F_{x_2}) \big)^2 $} 
\end{equation}
\noindent  and
\[ 
\mbox{\small $
F_{x_2}^2 \!+\! F_{y_2}^2  \leq  \mu^2 \!\cdot\! \big( \tfrac{1}{2} mg + \tfrac{h}{d} (F_{x_1} \!+\! F_{x_2}) \big)^2 . $} 
\]
\noindent When the normal load constraints of Eq.~\eqref{eq:normal} are taken into account, one can take the square-root of both sides of the 
non-sliding constraints
\[
\mbox{\small $
\sqrt{ F_{x_1}^2 \!+\! F_{y_1}^2 } \leq  \mu \!\cdot\!  \big( \tfrac{1}{2} mg - \tfrac{h}{d} (F_{x_1} \!+\! F_{x_2}) \big) $} 
\]
\noindent and
\[ 
\mbox{\small $
\sqrt{ F_{x_2}^2 \!+\! F_{y_2}^2 } \leq  \mu \!\cdot\! \big( \tfrac{1}{2} mg + \tfrac{h}{d} (F_{x_1} \!+\! F_{x_2}) \big) . $} 
\]
\noindent To express these constraints in terms of the control inputs, let us write the ground 
forces of Eq.~\eqref{eq:forces} as affine functions of the control inputs. The front wheel contact force~is~given~by
\[
\mbox{\small $
\begin{pmatrix}
F_{x_1} \\ F_{y_1}
\end{pmatrix}
= A_1(\nu,\phi) \! 
\begin{pmatrix} u_1 \\ u_2 \end{pmatrix} + \mbox{\boldmath ${b}$}_1(\nu,\phi) $} 
\] 
\noindent while the rear wheel contact force is given by
\[
\mbox{\small $
\begin{pmatrix}  F_{x_2} \\ F_{y_2} \end{pmatrix}
= A_2(\nu,\phi) \! 
\begin{pmatrix} u_1 \\ u_2 \end{pmatrix} + \mbox{\boldmath ${b}$} _2(\nu,\phi) $} 
\]
\noindent where expressions for the $2 \!\times\! 2$ matrices
$A_1$ and~$A_2$ and the vectors $\mbox{\boldmath ${b}$}_1$ and $\mbox{\boldmath ${b}$}_2$
are readily obtained from Eq.~\eqref{eq:forces}. Using the notation $\mbox{\boldmath ${u}$} \!=\! (u_1,u_2)$, the net $x$-axis ground force 
can be written as 
$F_{x_1} \!+\! F_{x_2} = \mbox{\boldmath ${a}$} \!\cdot\! \mbox{\boldmath ${u}$} + b$, 
where $\mbox{\boldmath ${a}$}$ is the sum of the upper rows of  $A_1$ and~$A_2$ and 
$b$ is the sum of the $x$-components of $\mbox{\boldmath ${b}$}_1$ and $\mbox{\boldmath ${b}$}_2$.
Using $\mbox{\boldmath ${u}$}$
and omitting the indices,~the~full non-sliding constraints take the form
\[
\mbox{\small $
\sqrt{\mbox{\small  $(A\mbox{\boldmath $u$} \!+\! \mbox{\boldmath $b$})^T (A\mbox{\boldmath $u$} \!+\! \mbox{\boldmath $b$})$} }
\pm \mu \tfrac{h}{d} \!\cdot\! ( \mbox{\boldmath $a$} \!\cdot\! \mbox{\boldmath $u$} \!+\! b)
\leq \tfrac{1}{2} \mu mg . $} 
\]
\noindent The first term on the left is the composition of the Euclidean norm with an affine function of~\mbox{\boldmath ${u}$}. Norms are convex functions and their composition with affine functions preserves convexity. The second term on the left is linear in \mbox{\boldmath ${u}$} and hence convex. The sum of convex functions forms a~convex function. The non-sliding constraints of
the bicycle model 
wheels thus form two convex sets in the controls $u_1$ and~$u_2$, see Fig.~\ref{fig:Hodograph} and \ref{fig:Hodograph0}.\\
\indent The {\em  full non-sliding constraints,} denoted $\mbox{\small $C$}_1$ and $\mbox{\small $C$}_2$, thus
take the form
\[
\mbox{\small $
 C_1(\nu,\phi,u_1,u_2) = \sqrt{ F_{x_1}^2 \!+\! F_{y_1}^2 } - \mu \!\cdot\!  \big( \tfrac{1}{2} mg - \tfrac{h}{d} (F_{x_1} \!+\! F_{x_2}) \big) \leq 0 $} 
\]
and
\[
\mbox{\small $
  C_2(\nu,\phi,u_1,u_2) = \sqrt{ F_{x_2}^2 \!+\! F_{y_2}^2 } -  \mu \!\cdot\! \big( \tfrac{1}{2} mg + \tfrac{h}{d} (F_{x_1} \!+\! F_{x_2}) \big) \leq 0 $} 
\]
where \mbox{\small $(F_{x_1},F_{y_1})$} and \mbox{\small $(F_{x_2},F_{y_2})$} are functions of the~state~var\-iables and controls. The allowed control set $\mathcal{U}$ is defined by the non-sliding constraints and the control limits
\[
\mbox{\small $
\begin{split}
\mathcal{U} = \big\{ 
(u_1,u_2)  : \,\, & \mbox{\small $C$}_1(\nu,\phi,u_1,u_2) \!\leq\! 0, 
\mbox{\small $C$}_2(\nu,\phi,u_1,u_2) \!\leq\! 0, \\
& |u_1| \!\leq\! a_{max}, |u_2| \!\leq\! b_{max} \big\}
\end{split} $} 
\]
\noindent forms a~{\em compact convex set}  with bounded 
state variables $|\nu| \!\leq\! \nu_{max}$~and~$|\phi| \!\leq\! \phi_{max}$. Under these conditions the car-like robot satisfies Filippov's existence theorem~\cite{filippov}: when a~path with valid controls exists between two endpoint states, there also exists a~time optimal path between these two states. Convexity of $\mathcal{U}$ also ensures that an~optimal control action can be uniquely determined at each time instant along the robot path. Note however that there can be several symmetrically arranged time-optimal paths for the same endpoint states (Fig.~\ref{fig:Example1traj}).

\section{Time Optimal Control Problem Analysis}

\noindent This section describes the {\em minimum principle} for the car-like robot 
time optimal control inputs under non-sliding constraints. The associated time optimal control problem is then solved for the path primitives that form the 
time optimal trajectories.

%

{\bf Minimum Principle:}
Consider the {\em augmented  Hamiltonian}~\cite{ben-asher,opt_survey} for the car-like robot. 
The Hamiltonian which consists of the robot state equations augmented by the state constraints 
takes the form
\[
\mbox{\small $ 
\begin{split}
H(t) = 	& \lambda _x(t) \nu(t)\cos \phi (t)\cos \theta (t) + \lambda _y(t) \nu(t)\cos \phi (t)\sin \theta (t) \\
&  + \tfrac{1}{d} \lambda _{\theta}(t) \nu(t)\sin \phi (t) + \lambda _v (t){u_1}(t) + {\lambda _\phi }(t) {u_2}(t) \\
& + \delta_\nu(t)  \big( \nu^2(t) \!-\! \nu^2_{max} \big) + \delta_\phi(t)  \big( \phi^2(t) \!-\! \phi^2_{max} \big) .
\end{split} $} 
\]
\noindent where {\small $(\lambda_x,\lambda_y,  \lambda_\theta, \lambda_{\nu},\lambda_\phi)$} are {\em costate variables}} \mbox{\hspace{-.31em} and \mbox{\hspace{-.36em}} $\delta_\nu(t) \!\geq\! 0$ } and $\delta_\phi(t) \!\geq\! 0$ are 
{\em state constraints} multipliers. The latter multipliers remain identically zero until the corresponding state constraint
becomes active, $|\nu| \!=\! \nu_{max}$ or $|\phi| \!=\! \phi_{max}$. Keeping this in mind, assume inactive state constraints with the Hamiltonian taking the simpler form
\begin{equation}  \label{eq:Hamiltonian} 
\mbox{\small $ 
\begin{split}
H(t) = 	& \lambda _x(t) \nu(t)\cos \phi (t)\cos \theta (t) + \lambda _y(t) \nu(t)\cos \phi (t)\sin \theta (t) \\
&  + \tfrac{1}{d} \lambda _{\theta}(t) \nu(t)\sin \phi (t) + \lambda _v (t){u_1}(t) + {\lambda _\phi }(t) {u_2}(t) 
\end{split} $} 
\end{equation}
\noindent where the vector of costate variables
\[
\mbox{\small $ 
\lambda(t) = \big( \lambda_x(t),\lambda_y(t),  \lambda_\theta(t), \lambda_{\nu}(t),\lambda_\phi(t) \big) $} 
\]
\noindent  is determined  by the 
{\em costate equations}~\cite{leitmann}[Sec. 13.10]
along time optimal paths
\[
\mbox{\small $ 
\begin{split}
\frac{d}{dt} \lambda(t) = & - \frac{\partial}{\partial \mbox{\small $S$}} H\big( \lambda(t), S(t), u_1(t), u_2(t) \big) \\[-2pt]
& - \mu_1(t) \frac{\partial}{\partial \mbox{\small $S$}} \mbox{\small $C$}_1(\nu,\phi, \mathbf{u})
- \mu_2(t) \frac{\partial}{\partial \mbox{\small $S$}} \mbox{\small $C$}_2(\nu,\phi, \mathbf{u})
\end{split} $} 
\]
\noindent where $\mbox{\small $S$} \!=\! ( \mbox{\small $X$}, \mbox{\small $Y$},\theta,\phi,\nu )$ and $\mathbf{u} \!=\! (u_1,u_2)$. Here,  $\mu_1(t)  \!\geq\! 0$ and $\mu_2(t)  \!\geq\! 0$ are \mbox{piecewise} \mbox{continuous} {\em \mbox{Lagrange} multipliers} of the wheels non-sliding constraints.  These multipliers remain identically zero until the corresponding non-sliding constraint
becomes active, $\mbox{\small $C$}_1(\nu,\phi,\mathbf{u}) \!=\! 0$ or $\mbox{\small $C$}_2(\nu,\phi,\mathbf{u}) \!=\! 0$.

Differentiating the simplified Hamiltonian~\eqref{eq:Hamiltonian} with respect to 
\mbox{\small $S$} gives the robot costate equations 
%
\[ 
\mbox{\small $ 
\begin{split}
\dot{\lambda}_x(t) & =  0 \,\, \Rightarrow \,\, \lambda_x = k_1 \\
\dot{\lambda}_y(t) & =  0 \,\, \Rightarrow \,\, \lambda_y = k_2 \\
\dot{\lambda}_{\theta}(t) & =   k_1 \nu(t) \sin\theta(t) \cos\phi(t) -  k_2 \nu(t) \cos\theta(t) \cos\phi(t) \\
\dot{\lambda}_{\nu}(t) & =  -k_1 \cos \theta(t) \cos\phi(t) - k_2 \sin\theta(t) \cos\phi(t) \\
& \hspace{-.1in} - \lambda_{\theta}(t) \tfrac{1}{d} \sin\phi(t) - \mu_1(t)   \frac{\partial C_1(\nu,\phi,\mathbf{u})}{\partial \nu} - \mu_2(t)   \frac{\partial C_2(\nu,\phi,\mathbf{u})}{\partial \nu}\\
\dot{\lambda}_\phi(t) & =   k_1 \nu(t) \cos\theta(t) \sin\phi(t) +
k_2 \nu(t) \sin\theta(t) \sin\phi(t) \\
& \hspace{-.1in} - \! \lambda_{\theta}(t) \nu(t)  \cos\phi(t)  \!-\! \mu_1(t)  \frac{\partial C_1(\nu,\phi,\mathbf{u})}{\partial \phi}  \!-\! \mu_2 (t) 
\frac{\partial C_2(\nu,\phi,\mathbf{u})}{\partial \phi} 
\end{split} $} 
\]
\noindent where $k_1$ and $k_2$ are constants.


To find candidate time optimal path segments, Pontryagin's {\em minimum principle}~\cite{pontryagin} is employed. At each time instant along the time optimal trajectory, the optimal control minimizes the Hamiltonian
\[ 
\mbox{\small $ 
\forall t \,\,  {H^*}(t) = \min_{u_1,u_2} \big\{ 
H\big( \mbox{\small $S$}(t),\lambda(t),  u_1(t), u_2(t) \big)
\big\} $} 
\]
\noindent where the minimization 
is subject~to~the~allow\-ed control set~$\mathcal{U}$.


%

\begin{table}[t]
\begin{center}
\vline
		\bgroup
		\def\arraystretch{2.1}
	\begin{tabular}{c|c|c|}
			\hline
			&	Front Wheels Linear & Steering Front Wheels\\[-8pt]
		 &	Acceleration $u_1(t)$ &	Rotation Rate $u_2(t)$ \\	
			\hline 
			${\cal L}^+$ &	${a_{max}}$ &	$b_{max}$ \\[-4pt]
			${{\cal L}^- }$ &	$-{a_{max}}$ &	$b_{max}$ \\[-4pt]
			${{\cal L}^0 }$ &	$0$         &	${b_{max }}$\\[-4pt]
			${{\cal R}^+ }$ &	${a_{max}}$	&   $-b_{max }$\\[-4pt]
			${{\cal R}^- }$ &	$-{a_{max}}$	& $-b_{max}$\\[-4pt]
			${{\cal R}^0 }$ &	$0$         & $-b_{max}$ \\[-4pt]
			${\cal S}^+$ & ${a_{max}}$ & $b_{singular}(t)$\\[-4pt]
		${{\cal S}^- }$ &	${-a_{max}}$ & $b_{singular}(t)$\\[-4pt]
			${{\cal S}^0 }$ &	$0$    & $b_{singular}(t)$\\[-4pt]
		${{\cal C}^+}$ &	${a_{max}}$	& $  0$ \\[-4pt]
			${{\cal C}^- }$ &	${-a_{max}}$	& $  0$ \\[-4pt]
			${{\cal C}^0 }$ &	$0$         &	$  0$  \\
			\hline
		\end{tabular}
		\egroup 
	\end{center}
	\caption{The twelve time optimal path primitives of the car-like robot moving under inactive non-sliding constraints.}
	\label{tab:PathPrimitives}
\end{table}

{\bf Time optimal path primitives} For the kinematic car-like robot without the non-sliding constraints,  Ref.~\cite{yossi_car} obtained {\em twelve path primitives} listed in Table~I.
These  primitives form the possible time-optimal path segments under inactive non-sliding constraints divided 
into three types.
The first type are  {\em regular controls,}
extremal controls
determined directly from the minimum principle: 
${\cal L}^+$, ${\cal L}^-$, ${\cal L}^0$ and ${\cal R}^+$,  ${\cal R}^-$, ${\cal R}^0$. 
The time optimal 
control inputs along these path segments are at their respective limits (bang-bang controls), with $u_1 \!=\! 0$ in the case of ${\cal L}^0$ and ${\cal R}^0$. 
The second type are {\em singular controls,} determined by the vanishing of a switching function and its time derivatives: ${\cal S}^+$, ${\cal S}^-$ and ${\cal S}^0$. The third type are associated with active state constraints: ${\cal C}^+$, ${\cal C}^-$ and ${\cal C}^0$, 
where the time-optimal 
control inputs are either at their limits or zero when the corresponding state, speed~$\nu$ or steering angle~$\phi$, reaches its limit. While $u_1 \!=\! \dot{\nu}$ is always regular,
a~singular steering-rate control $u_2 \!=\!  \dot{\phi}$ exists along the ${\cal S}^+$, ${\cal S}^-$ and ${\cal S}^0$ primitives. It is given by
\[ 
\mbox{\small $ 
u_2(t) = b_{singular}(t) =  \nu(t) \cos\phi(t)  \tan(\theta(t) \!-\! \theta_0)/d. $} 
\]
\noindent where the 
costate variables $\lambda_x \!=\! k_1$ and $\lambda_y \!=\! k_2$ determine the 
car-like robot heading parameter $\theta_0$:
$k_1 \sin\theta_0  \!-\! k_2 \cos\theta_0 \!=\! 0$.

Next consider the car-like robot model when the non-sliding constraints become active under the controls~$u_1 \!=\! \dot{\nu}$~and $u_2 \!=\! \dot{\phi}$ while
$u_1$ and $u_2$ lie inside their limits,
$|u_1| \!\leq\! a_{max}$ and $|u_2| \!\leq \! b_{max}$. 
The key analysis tool here is called  the {\em Hodograph} of the controls.
The Hodograph is a~snapshot of the allowed control set~$\mathcal{U}$ at some speed and steering angle as seen in Figs.~\ref{fig:Hodograph} and~\ref{fig:Hodograph0}.
Minimization of the Hamiltonian from Eq.~\eqref{eq:Hamiltonian} 
requires minimization of $\lambda_\nu(t) u_1 \!+\!\lambda_\phi(t) u_2$ subject to the control constraints. 
Since the Hamiltonian is linear in $u_1$ and $u_2$, 
the optimal controls occur at the {\em maximal projection} of the control set~$\mathcal{U}$ along the costate vector $-(\lambda_\nu(t), \lambda_\phi(t))$ ($-\lambda$ in Figs.~\ref{fig:Hodograph}-\ref{fig:Hodograph0}). The car-like robot time optimal trajectories thus consist of twelve path primitives with inactive sliding constraints and three 
path primitives where
either one or both wheels are on the verge of sliding.

\begin{table}[t]
\begin{center}
\vline
	\bgroup
		\def\arraystretch{2.1}
	\begin{tabular}{c|c|c|}
		\hline
     	&	Front Wheels & Rear Wheels \\[-8pt]
            &   Ground Force & Ground Force \\
		\hline
            ${\cal FW}$ & $\left\| F_1 \right\|^2 = \mu^2  (N_1)^2 $ & 
            --- \\[-4pt]
			${\cal RW}$ & ---  & $\left\| F_2 \right\|^2 = \mu^2 (N_2)^2 $   \\[-4pt]
		${\cal BW}$ & $\left\| F_1 \right\|^2 = \mu^2  (N_1)^2 $ &  $\left\| F_2 \right\|^2 = \mu^2 (N_2)^2 $ \\
		\hline		
  \end{tabular}
		\egroup 
	\end{center}
\caption{The time optimal path primitives of the car-like robot moving under~\mbox{active}~\mbox{non-sliding}~\mbox{constraints.}~\mbox{Contact}~forces~and~normal loads are state dependent and linear in~$u_1$ and $u_2$.}
	\label{tab:PathPrimitives1}
\end{table}

Table~II lists the time optimal path primitives associated with active non-sliding constraints: $\cal {FW}$ when the forward wheel is on the verge of sliding, $\cal {RW}$ when the rear wheel is on the verge of sliding, $\cal {BW}$ when both wheels are on the verge of sliding.  Solutions for the $\cal {FW}$ and $\cal {RW}$ primitives can only be obtained numerically by solving the 
state and costate equations.
Approximate analytic solutions for these primitives  
are described in the paper appendix. In the appendix,  each primitive $\cal {FW}$ or $\cal {RW}$ is clipped by the control limits  $u_1 \!=\! \pm a_{max}$ and $u_2 \!=\! \pm b_{max}$. This gives up to four segments of the non-sliding constraint for each wheel that can possibly contain the time optimal solution according to the minimum principle. The endpoints of each segment are taken as approximate solution pairs for the $\cal {FW}$ and~$\cal {RW}$ primitives, with excellent approximation quality discussed in the appendix.

%



{\bf Dominance of \boldmath{$\cal {FW}$} over \boldmath{$\cal {RW}$}:} The front wheel~non-sliding constraint is active
more often than the~rear~wheel~non-sliding constraint along time optimal paths. To observe this property,
assume 
that the bicycle-robot  center of mass is sufficiently low so that dynamic effects on the wheels vertical loads at the ground contacts  can be neglected. The vertical load on each wheel is simply $\tfrac{1}{2} mg$. Under this assumption the determinants of the contact force matrices,  $A_1$ for $\cal {FW}$ and $A_2$ $\cal {RW}$ from Section~IV, are given by
\[
\det (A_1) = \tfrac{1}{4} (m \!+\! m_w) (\mbox{\small $1$} \!+\! \mbox{\small $4$} \tfrac{\rho^2}{d^2} ) \!\cdot\! \nu  
\]
\noindent and 
\[
\det (A_2) = - \tfrac{1}{4} m_w  (\mbox{\small $1$} \!-\! \mbox{\small $4$} \tfrac{\rho^2}{d^2} ) \!\cdot\! \nu  
\]
\noindent where the bicycle robot mass~$m$ has been canceled out.
The areas of the front and rear wheels non-sliding ellipses, $\pi/|\det(A_1)|$ and $\pi/ |\det(A_2)|$, are bounded by the ratio $|\det(A_2)| / |\det(A_1)| \!\leq\! m_w / m$. As each bicycle wheel mass, $2m_w$, 
is much smaller than the bicycle robot total mass~$m$, the $\cal {FW}$ primitive ellipse is much smaller and hence dominates the $\cal {RW}$ primitive along time optimal paths (Fig.~\ref{fig:Hodograph}). The $\cal {RW}$ primitive typically occurs along high curvature path segments due to dynamic vertical loads effects  (Fig.~\ref{fig:traj4}). One can also verify that the non-sliding constraint of each wheel forms a~vertical strip in the $(u_1,u_2)$ plane at zero speed, then a~convex bounded region (usually an~ellipse) that shrinks in size as~the~bicycle~robot~speed increases. 

\begin{figure}
\centering	
\includegraphics[width=0.65\linewidth]{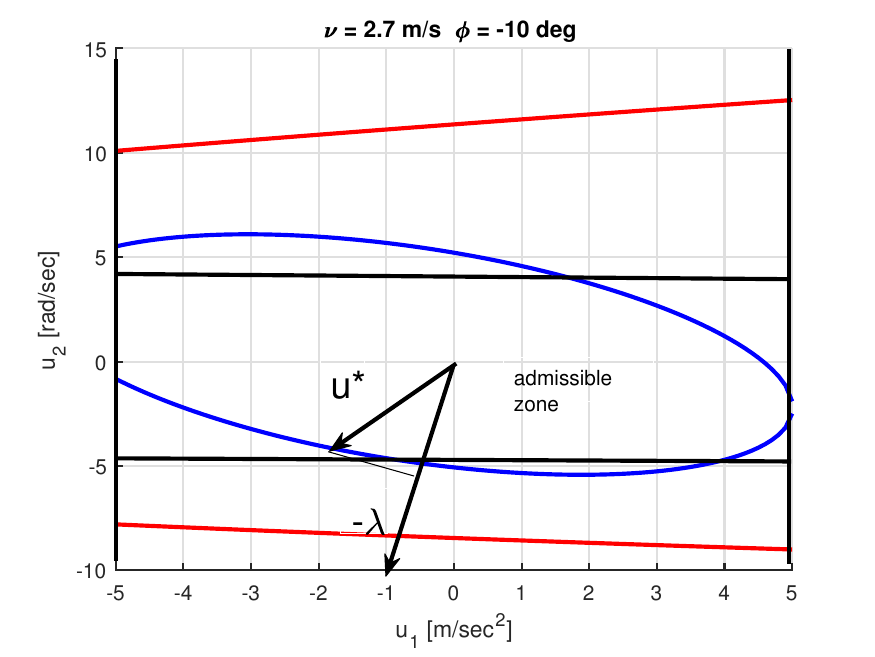}
\caption{Hodograph of the path primitive $\cal {FW}$ with active front wheel non-sliding constraint  (blue curve) and inactive rear wheel non-sliding constraint (red curve).
Horizontal and vertical black lines are inactive control limits.
Hodograph is captured at $t \!=\! 1.0$~$\mathrm{sec}$ 
during  $\cal {FW}$ of Example~1 in Section~\ref{sec.sim}. }
	\label{fig:Hodograph}
\end{figure}

\begin{figure}
\centering
\includegraphics[width=0.65\linewidth]{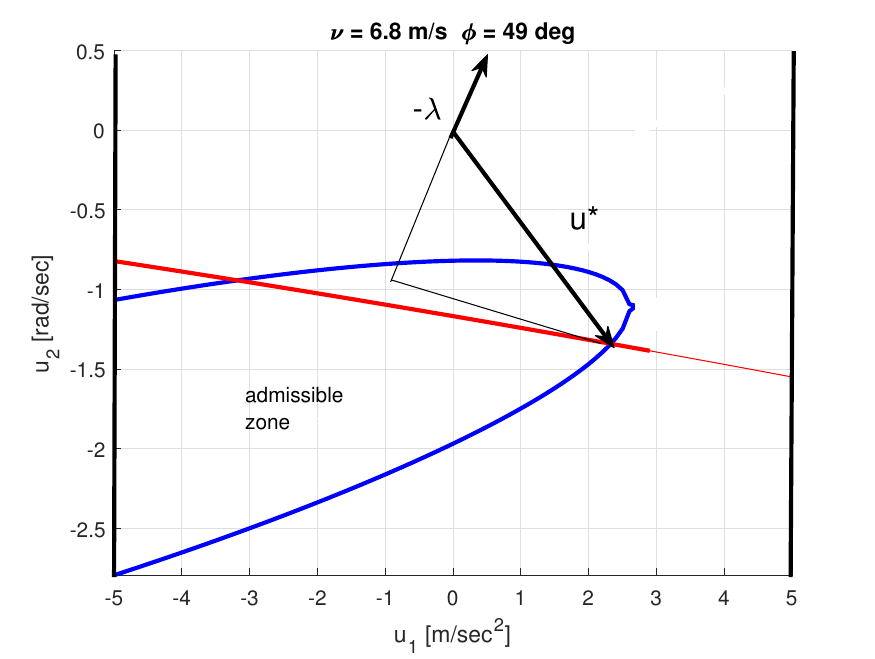}
\caption{Hodograph of the path primitive $\cal {BW}$ with both front and rear wheel non-sliding constraints active  (front wheel blue curve, rear wheel red curve). Vertical black lines are inactive control limits. Hodograph captured at $t \!=\! 2.3$~$\mathrm{sec}$ 
during  $\cal {BW}$ of Example~3 in Section~\ref{sec.sim}.}
	\label{fig:Hodograph0}
\end{figure} 

\section{Representative Examples}  \label{sec.sim}

\noindent This section describes numerical examples that demonstrate and validate the analysis. For these examples, the control and state limits are $a_{max} \!=\! 5$~$\mathrm{m / sec^2}$, $b_{max} \!=\!  3\pi / 2$~$\mathrm{rad/sec}$,
$\nu_{max} \!=\! 20$~$\mathrm{m/sec}$ and $\phi_{max} \!=\! 90^{\circ}$. The car-like robot mass is $m \!=\! 20$~$\mathrm{kg}$, its length is $d \!=\! 2$~$\mathrm{m}$ and its center of mass height is set at $h \!=\! 0.2$~$\mathrm{m}$.
The rear wheel mass is taken as $m_w \!=\! 1$~$\mathrm{kg}$ with radius $r_w \!=\! 0.1$~$\mathrm{m}$. The examples assume a~uniform coefficient of friction $\mu \!=\! 1.0$ with the last two examples comparing $\mu \!=\! 1.0$ against using $\mu \!=\! 2.0$. 

Solutions for the examples were obtained by direct optimization using the optimal control package {\small GPOPS-II}~\cite{gpops2}. It approximates the optimal control problem  as a~discrete nonlinear optimization problem solved by user-selected techniques such as {\small IPOPT} or {\small SNOPT}~\cite{ben-asher}. Execution times of the examples were $3.0$-$3.7$~$\mathrm{sec}$ using {\small SNOPT} and $3.6$-$5.4$~$\mathrm{sec}$ using {\small IPOPT} on
Intel Core  $1.99$~$\mathrm{GHz}$ {\small I7-8550U CPU}.
Once the time optimal solution of the nonlinear optimization problem has been computed, costate estimates can be directly obtained from the Lagrange multipliers associated with each equality constraint~\cite{ben-asher}. The controls and costates are then used to identify the 
time optimal primitives and test for the minimum principle satisfaction on representative Hodographs.

\begin{figure}
\centering
\includegraphics[width=0.65\linewidth]{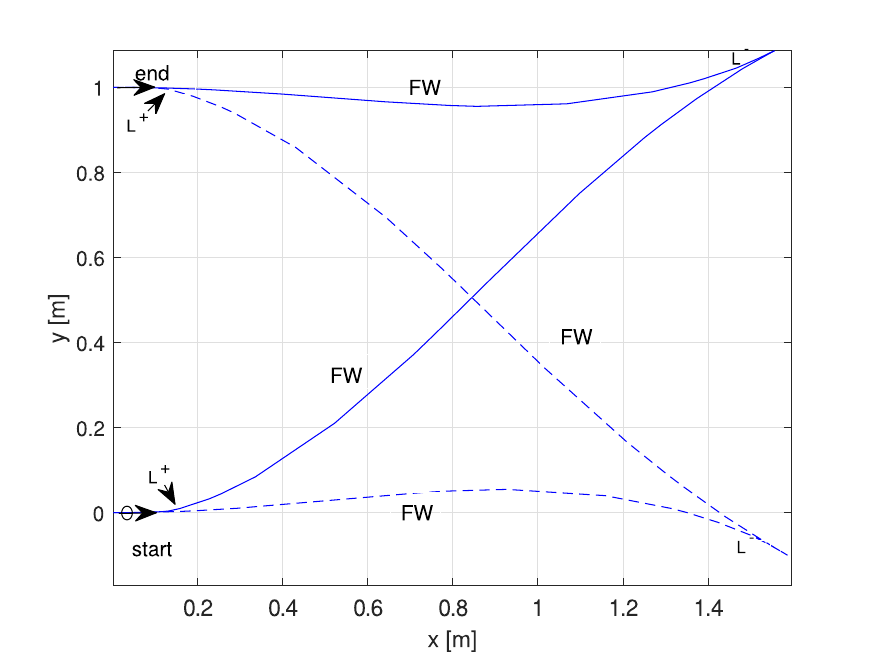}
\caption{The time optimal path for the parallel parking maneuver of Example~1. The car-like robot reaches the cusp with forward motion at $t \!=\! 1.4$~$\mathrm{sec}$, then moves backward while maintaining continuous path curvature. An alternative time optimal parallel parking maneuver (dashed curve) follows symmetrically arranged forward and backward motions (see also video clip).}
	\label{fig:Example1traj}
\end{figure}

\begin{figure}
	\centering
	\includegraphics[width=0.65\linewidth]{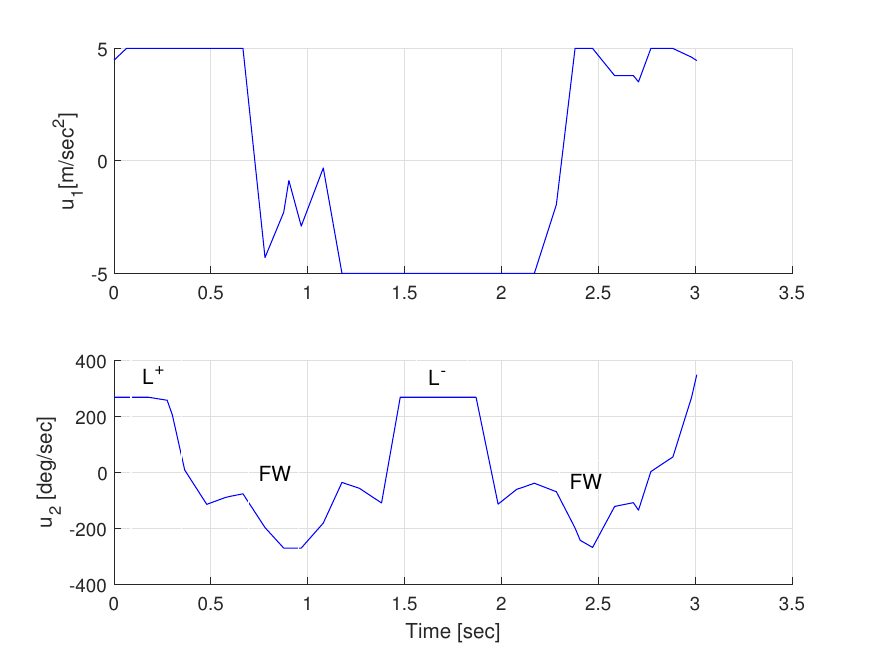}
\caption{The control inputs $u_1$ and $u_2$ for Example~1.}
	\label{fig:Example1Control}
\end{figure}

\indent {\bf Example 1:} This example describes 
a~parallel~parking maneuver~of~the car-like  robot (see also video clip). Using state coordinates $\mbox{\small $S$} \!=\! ( \mbox{\small $X$}, \mbox{\small $Y$},\theta,\nu,\phi )$ the example considers the endpoint states 
\mbox{\small $ 
\hspace{-.05in} S(0) = ( 0~\mathrm{m}, 0~\mathrm{m}, 0^{\circ}, 0~\mathrm{\tfrac{m}{sec}}, 0^{\circ} ) $} 
and
\mbox{\small $
S(t_f) = (0~\mathrm{m}, 1~\mathrm{m},0^{\circ}, 0~\mathrm{\tfrac{m}{sec}}, 0^{\circ} )$.} 
The time optimal path depicted in Fig.~\ref{fig:Example1traj} consists of the path primitives ${\cal L}^+$,  ${\cal FW}$, ${\cal L}^-$ and $\cal {FW}$. These primitives can be verified by inspecting the control inputs of Fig.~\ref{fig:Example1Control}. When the control input $u_1 \!=\! \dot{\nu}$  takes intermediate~non-constant~val\-ues, it results from an~active non-sliding constraint ($\cal {FW}$ in Fig.~\ref{fig:Example1traj}). 
To validate the minimum principle, a~snapshot of the Hodograph taken
at $t \!=\! 1.0$~$\mathrm{sec}$  
during  ${\cal FW}$ is shown in Fig.~\ref{fig:Hodograph}.
Note how the optimal control $\mathbf{u}^* \!\in\! \mathcal{U}$ maximizes its projection along~$-\lambda$~at~this~time~\mbox{instant.}~\eex
%

\begin{figure}
\centering
\includegraphics[width=0.65\linewidth]{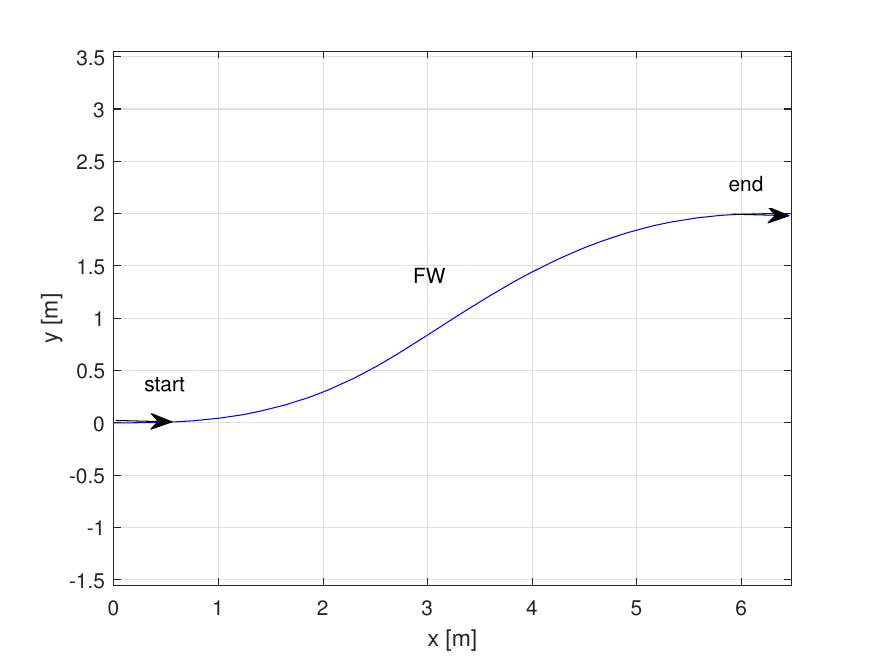}
\caption{The time optimal path for the lane transition maneuver of Example~2,
compare to the higher ground friction lane transition maneuver of Example~5.} \label{fig:traj2}
\end{figure}

\begin{figure} 
\centering
\includegraphics[width=0.65\linewidth]{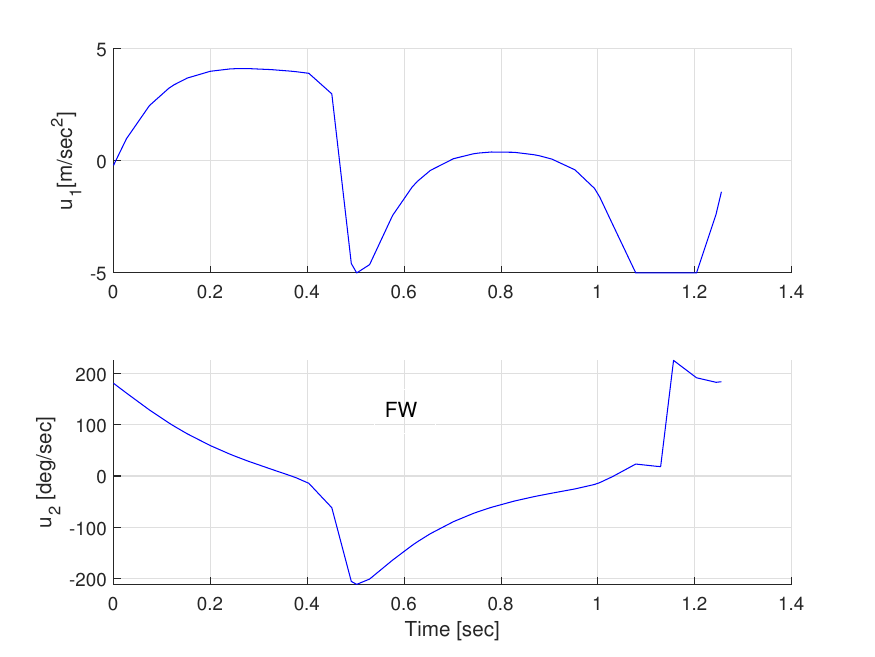}
\caption{The control inputs $u_1$ and $u_2$ for Example~2.} \label{fig:control2}
\end{figure}

\begin{figure}
	\centering
\includegraphics[width=0.65\linewidth]{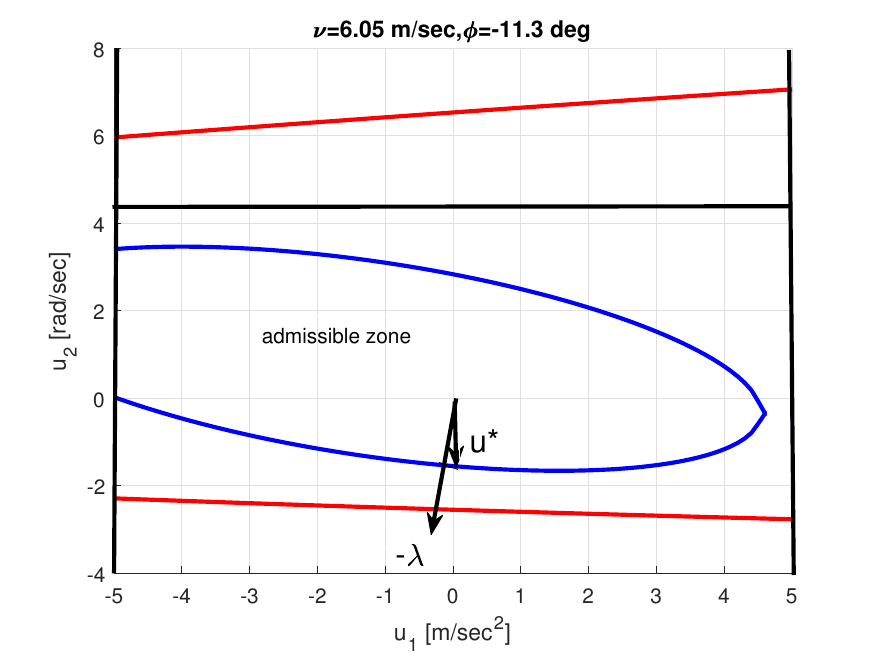}
\caption{The Hodograph at $t \!=\! 0.7$~$\mathrm{sec}$ during the path primitive $\cal {FW}$ of Example~2
(front wheel blue curve, rear wheel red curve). Black lines are inactive control limits.}
\label{fig:Example2hod}
\end{figure}

{\bf Example 2:} This example describes 
a~lane transition maneuver of the car-like  robot using high endpoint spee. 
Consider the endpoint states
\mbox{\small $
S(0) = ( 0~\mathrm{m}, 0~\mathrm{m}, 0^{\circ}, 5~\mathrm{\tfrac{m}{sec}}, 0^{\circ} ) $} 
and
\mbox{\small $
S(t_f) = (x_f, 2~\mathrm{m},0^{\circ}, 5~\mathrm{\tfrac{m}{sec}}, 0^{\circ} ) $} 
where $x_f$ is kept as a~free parameter for the lane transition maneuver. 
The time optimal path depicted in Fig.~\ref{fig:traj2} consists of the single primitive, ${\cal FW}$, associated with active front wheel non-sliding constraint. This primitive can be verified by inspecting the  controls along this path shown in  Fig.~\ref{fig:control2}.
To validate the minimum principle, a~snapshot of the Hodograph at $t \!=\! 0.7$~$\mathrm{sec}$ is shown in Fig.~\ref{fig:Example2hod}. Again, the optimal control $\mathbf{u}^* \!\in\! \mathcal{U}$
maximizes its projection along $-\lambda$ at this time instant.
Note that $u_1 \!=\! a_{max}$ towards the end of the ${\cal FW}$ primitive.~\eex
%

\newpage

\begin{figure} 
\centering
\includegraphics[width=0.65\linewidth]{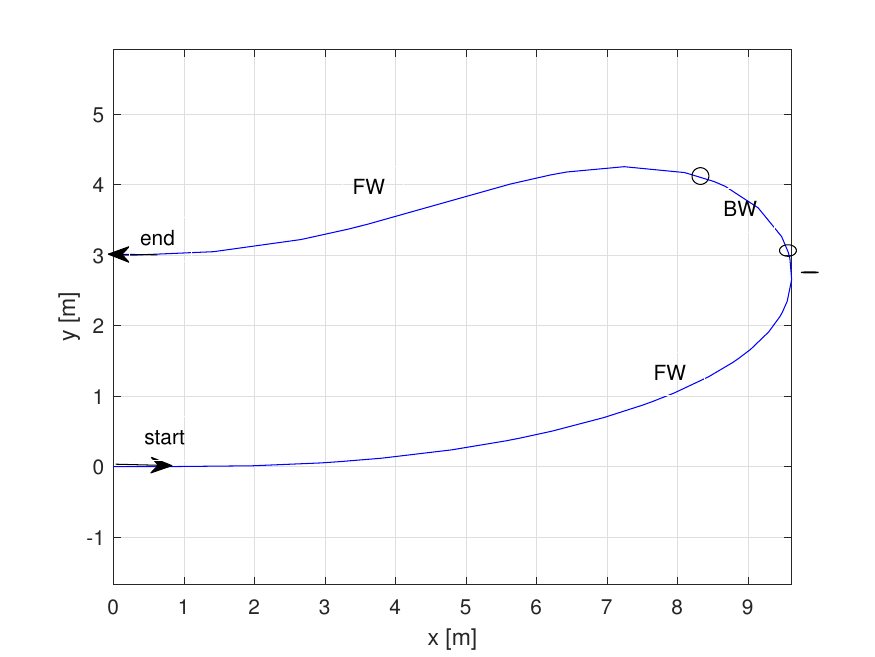}
\caption{The time optimal path for the U-turn maneuver of Example~3.}
\label{fig:traj3}
\end{figure}

\begin{figure}
\centering
\includegraphics[width=0.65\linewidth]{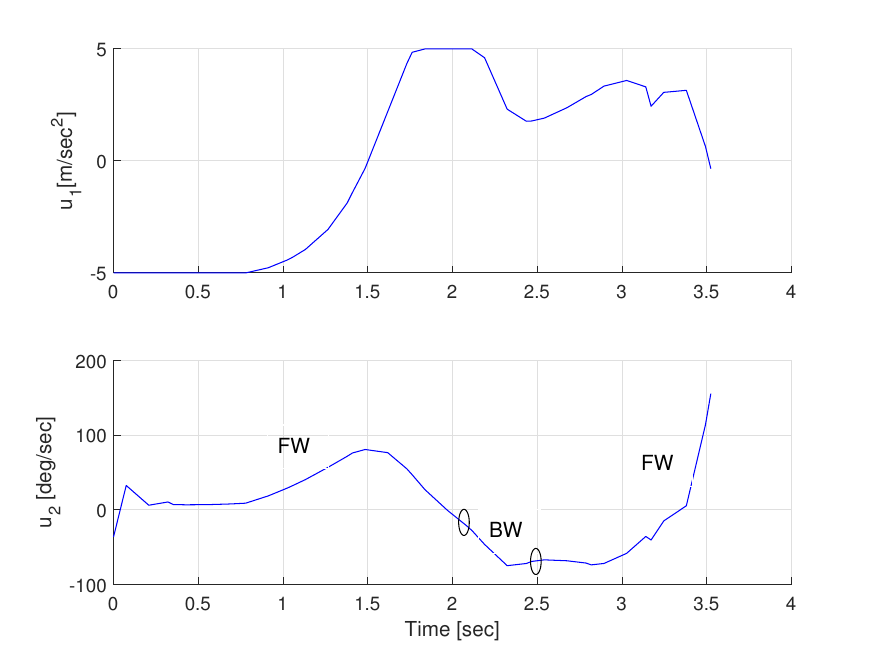}
\caption{The control inputs $u_1$ and $u_2$ for Example 3.}
\label{fig:control3}
\end{figure}

{\bf Example 3:} This example describes 
a~{\small U}-turn maneuver of the car-like robot.
Using state coordinates $\mbox{\small $S$} \!=\! ( \mbox{\small $X$}, \mbox{\small $Y$},\theta,\nu,\phi )$ consider the endpoint states
\mbox{\small $
\hspace{-.06in} S(0) = ( 0~\mathrm{m}, 0~\mathrm{m}, 0^{\circ}, 10~\mathrm{\tfrac{m}{sec}}, 0^{\circ} ) $} 
%
and
\mbox{\small $
S(t_f) = (0~\mathrm{m}, 3~\mathrm{m},180^{\circ}, 10~\mathrm{\tfrac{m}{sec}}, 0^{\circ} ). $} 
The time optimal path depicted in Fig.~\ref{fig:traj3} consists 
of the primitives ${\cal FW}$, ${\cal BW}$ and ${\cal FW}$.
Fig.~\ref{fig:control3} shows the controls along this path, showing that ${\cal FW}$ dominates most of this maneuver. Immediately after turning back,  the rear wheel non-sliding constraint also becomes active for a~brief period of $0.4$ seconds  (Fig.~\ref{fig:traj3}). Thus we encounter the path primitive $\cal {BW}$. Towards the end, $\cal {FW}$ is once again the only active 
constraint.\\
\indent Simultaneous activation of the non-sliding constraints on both wheels is known to be {\em intentionally violated} in drift car competitions. It is clearly shown on the corresponding Hodograph shown in Fig.~\ref{fig:Hodograph0}.
The recurrent fact that the optimal control $\mathbf{u}^* \!\in\! \mathcal{U}$ maximizes its projection along~$-\lambda$~is a validation of the minimum principle.~\eex
%

\begin{figure}
\centering
\includegraphics[width=0.65\linewidth]{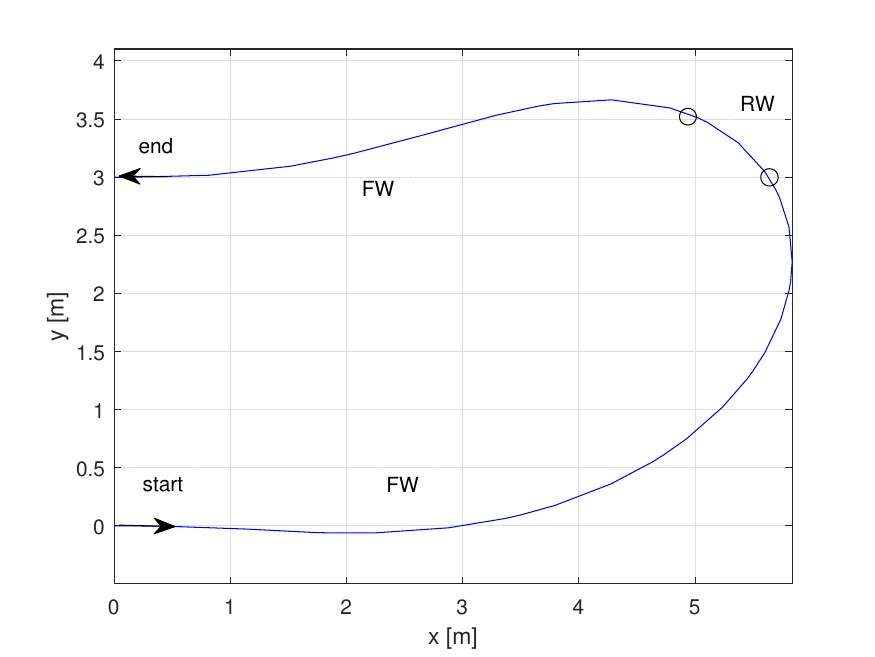}
\caption{The time optimal path for the high friction U-turn maneuver of Example~4. Note the $\cal {RW}$ path primitive along highest curvature segment.} 
\label{fig:traj4}
\end{figure}

\begin{figure} 
\centering
 \includegraphics[width=0.65\linewidth]{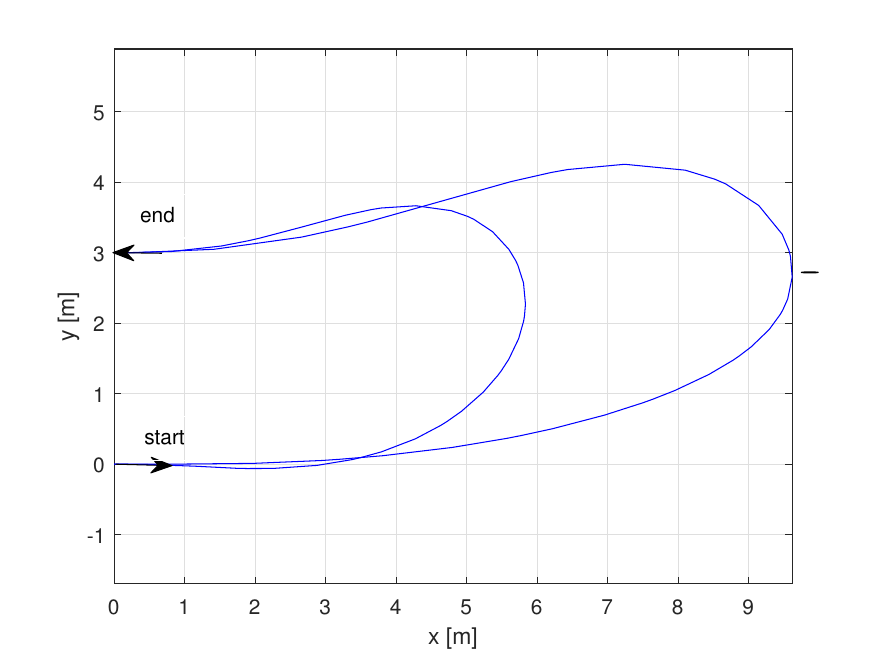}
 \caption{The time optimal paths for the U-turn maneuvers of Example~3 (outer path,  $\mu \!=\! 1.0$) and Example~4 (inner path, $\mu \!=\! 2.0$).} \label{fig:traj34}
\end{figure}

\begin{figure} 
\centering
\includegraphics[width=0.65\linewidth]{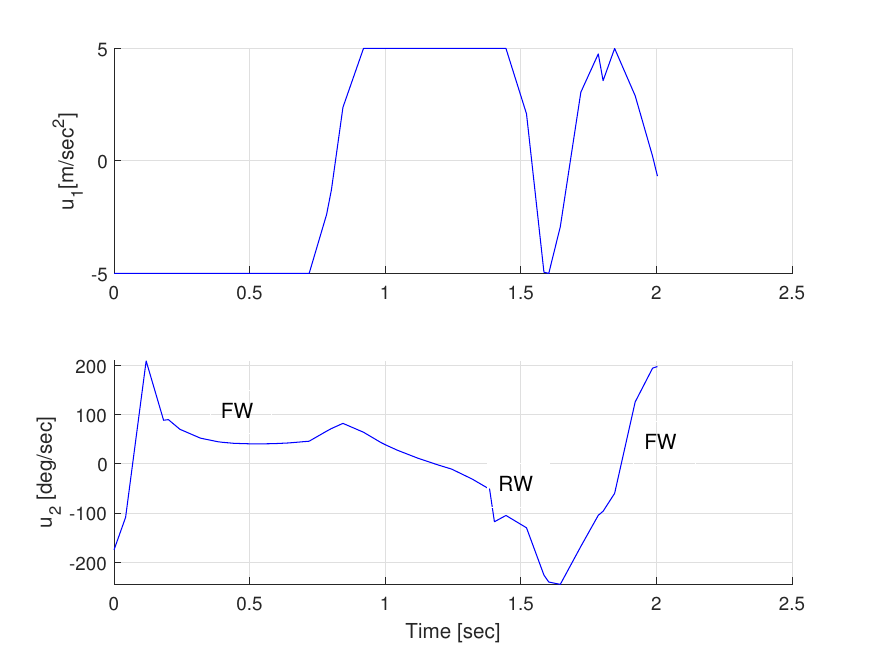}
\caption{The control inputs $u_1$ and $u_2$ for Example 4.}
\label{fig:control4}
\end{figure}

\begin{figure}
\centering	\includegraphics[width=0.65\linewidth]{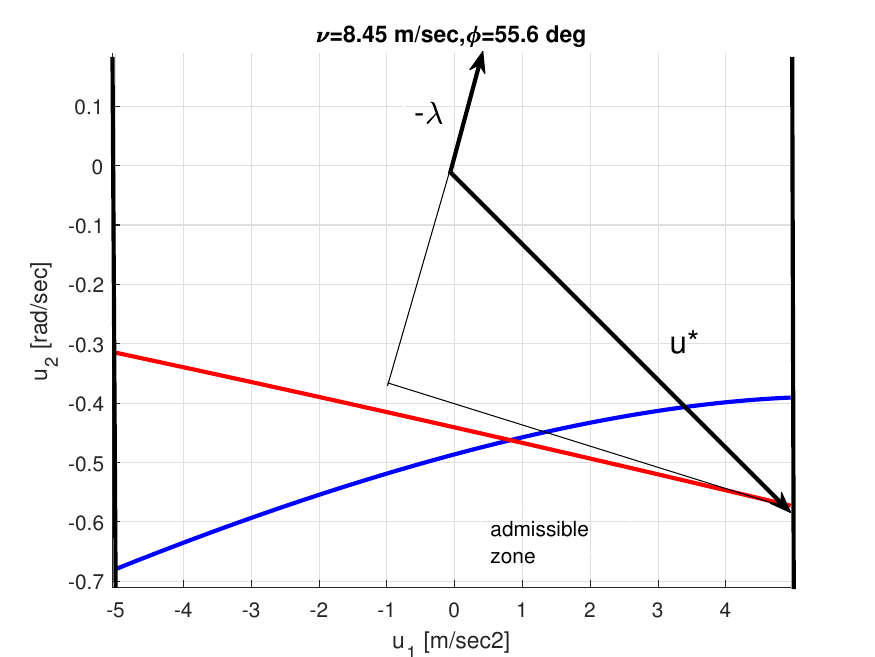}
\caption{The Hodograph at $t \!=\! 1.3$~$\mathrm{sec}$ during the path primitive $\cal {RW}$  of Example~4 (front wheel blue curve, rear wheel red curve).}
\label{fig:Hodograph.Ex4}
\end{figure}

{\bf Example 4:} This example uses a~higher coefficient of friction $\mu \!=\! 2.0$ for the {\small U}-turn maneuver.  The  time optimal path under higher $\mu$ is depicted Fig.~\ref{fig:traj4}. The $\mu \!=\! 1.0$ and $\mu \!=\! 2.0$ time optimal paths are shown together in Fig.~\ref{fig:traj34}. The travel time has decreased from $3.1$~$\mathrm{sec}$  to $2.0$~$\mathrm{sec}$.
Fig.~\ref{fig:control4} shows the controls $u_1$ and $u_2$ along this path. The time optimal path primitives under $\mu \!=\! 2.0$
are now $\cal {FW}$, $\cal {RW}$ and $\cal {FW}$. Immediately after turning back,  the rear wheel non-sliding constraint  becomes active for a brief period of $0.15$~seconds while the front wheel constraint becomes inactive. Thus we encounter the path primitive $\cal {RW}$ where only the rear wheel is on the verge of sliding. Towards the end, $\cal {FW}$ is once again the only active constraint. The  Hodograph shown in Fig.~\ref{fig:Hodograph.Ex4} at $t \!=\! 1.3$~$\mathrm{sec}$ shows the path primitive $\cal {RW}$. Again, the optimal control $\mathbf{u}^*  \!\in\! \mathcal{U}$ maximizes its projection along $-\lambda$ as required by the minimum principle.~\eex

\begin{figure}
\centering
\includegraphics[width=0.65\linewidth]{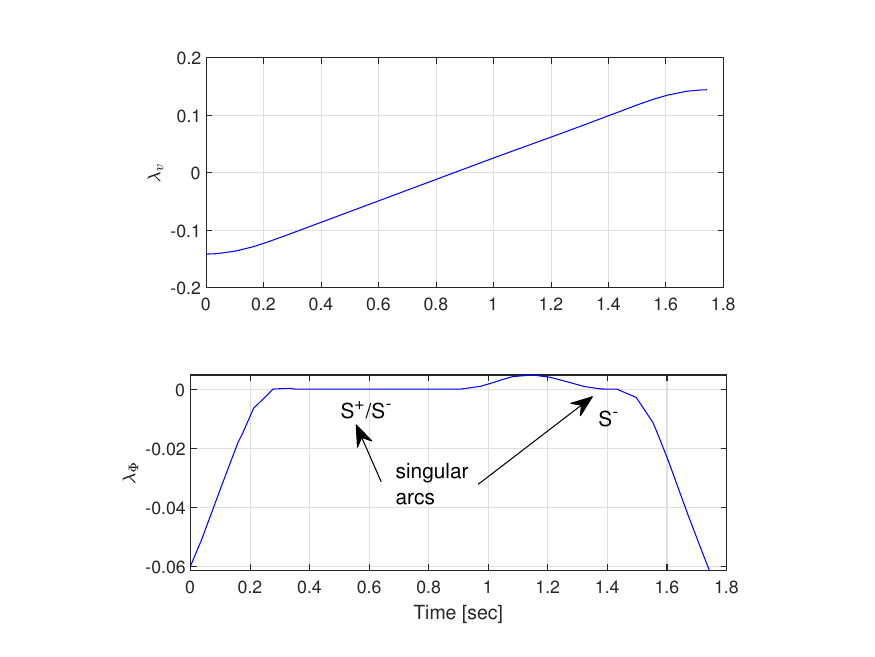}
\caption{The costate $\lambda_{\nu}$ and the singular costate $\lambda_{\phi}$ along the time optimal path of Example~5. Note that $\lambda_{\phi} \!=\! 0$ along the singular~path~primitives.} \label{fig:costate5}
\end{figure}

\begin{figure} 
\centering
\includegraphics[width=0.65\linewidth]{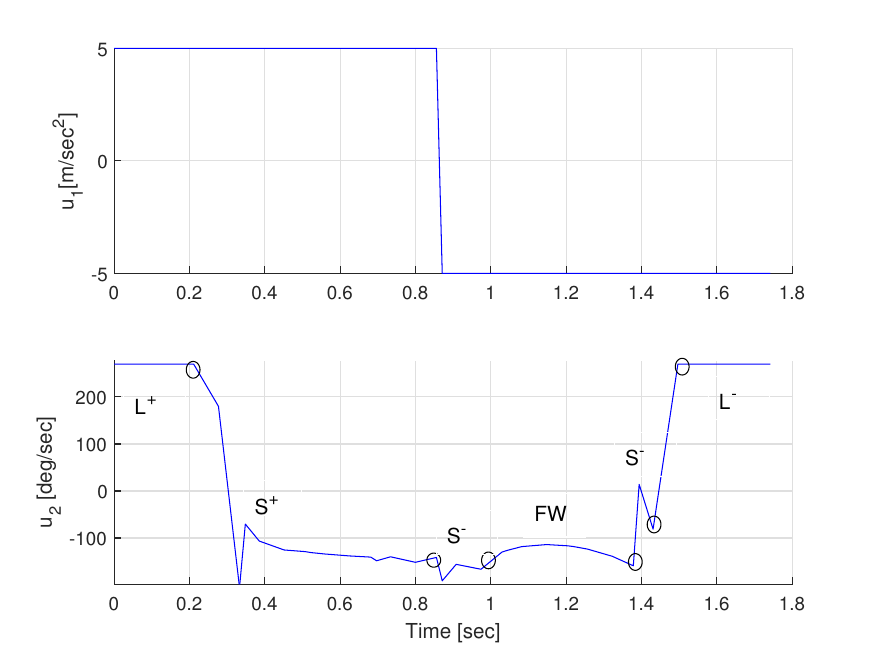}
\caption{The control inputs $u_1$ and $u_2$ for Example 5.} 
\end{figure}

\begin{figure}
	\centering
\includegraphics[width=0.65\linewidth]{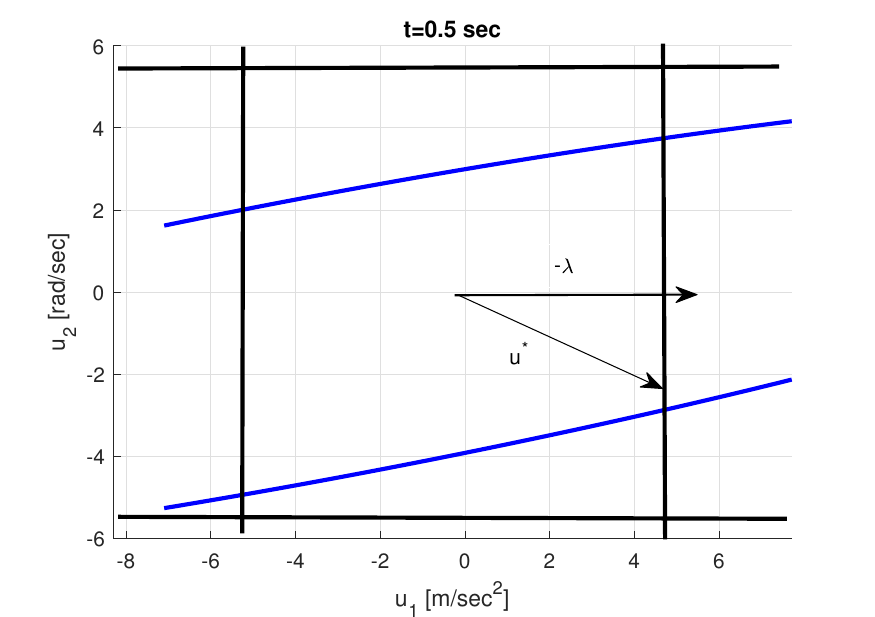}
\includegraphics[width=0.65\linewidth]{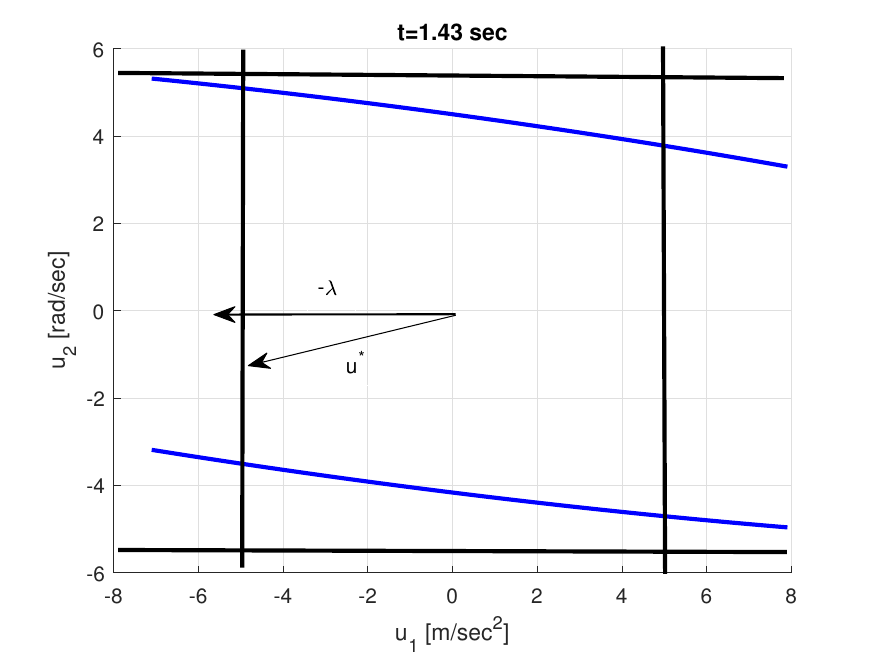}
\caption{The Hodographs at $t \!=\! 0.5$~$\mathrm{sec}$ and  $t \!=\! 1.43$~$\mathrm{sec}$
during the path primitive $\cal {FW}$ of Example~5
(front wheel blue curve, rear wheel red curve). Black lines are inactive control limits. }
\label{fig:Example6hod}
\end{figure} 

\begin{figure} 
\centering
\includegraphics[width=0.65\linewidth]{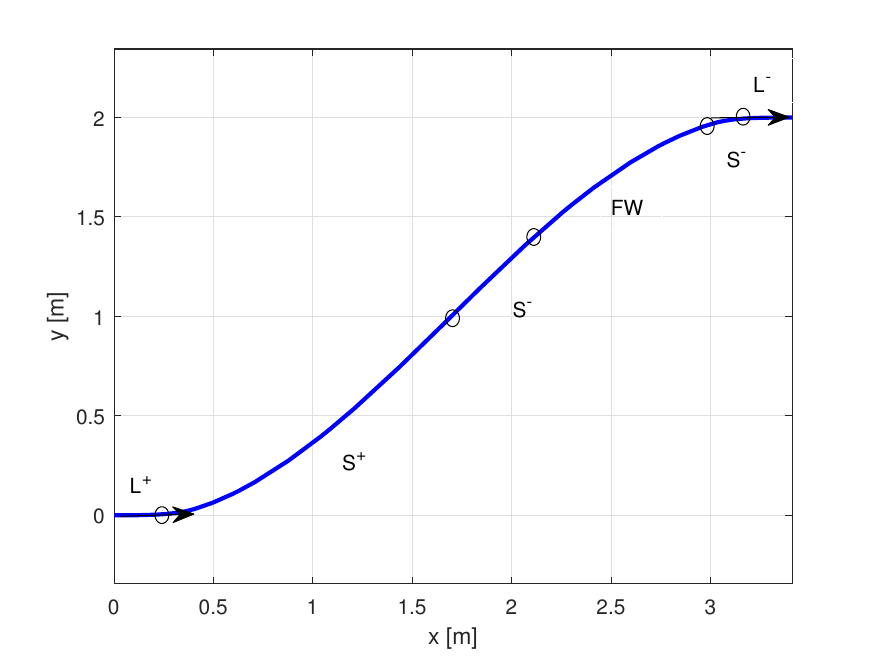}
\caption{The time optimal path for the lane transition maneuver of Example~5, showing the path primitives from Table~I with a single ${\cal FW}$ path primitive. Compare to Example~2 where ${\cal FW}$ dominates the entire path} \label{fig:traj5}
\end{figure}

{\bf Example 5:} This example shows 
path primitives associated with {\em inactive} non-sliding constraints. The lane transition maneuver is repeated with center of mass at a~lower height of $h \!=\! 0.1$~$\mathrm{m}$, coefficient of friction $\mu \!=\! 2.0$ and lower endpoint speeds of 
$\nu \!=\! 1$~$\mathrm{m/sec}$. The time optimal path depicted Fig.~\ref{fig:traj5} starts with ${\cal L}^+$ 
followed by the singular  ${\cal S}^+$  primitive associated with 
$u_1 (t) \!=\! a_{max}$ and $u_2(t) \!=\! b_{singular}(t)$.
During the ${\cal S}^+$ path primitive, the 
non-sliding constraint ${\cal FW}$ becomes 
active at the single time~instant~$t \!=\! 0.33$~$\mathrm{sec}$ located at $(x,y) \!=\! (0.33,0.02)$. Such single time instant activation of two path primitives is known as a~{\em touch point.} The time optimal path continues with the ${\cal S}^+$ primitive
to the path midpoint where it switches to the ${\cal S}^-$ primitive associated with 
$u_1 (t) \!=\! -a_{max}$ and $u_2(t) \!=\! b_{singular}(t)$.
The ${\cal FW}$ primitive appears at the end of  ${\cal S}^-$ followed by final ${\cal S}^-$ and ${\cal L}^+$ primitives. Note that the singular path primitives asymptotically converge towards or diverge away from straight line motions with  $\lambda_{\phi} \!=\! 0$ along these primitives (Fig.~\ref{fig:costate5}).~\eex 

%
%

%
%
%

%
%


\section{Conclusion}

\noindent The paper analyzed the car-like robot non-sliding constraints within the relatively simple full kinematic model of this robot. The car-like robot full kinematic model was augmented with dynamic based expressions for the ground 
forces using the bicycle model simplification. The ground 
forces are affine in the robot control inputs $u_1$ and $u_2$, a property that implies convexity of the non-sliding constraints and hence convexity of the allowed control set. The Hodograph method was then used to
augment the twelve time optimal path primitives of the car-like robot with three additional time optimal path primitives associated with the non-sliding constraints. These
are  $\cal {FW}$, $\cal {RW}$ and $\cal {BW}$ when the car-like robot moves with its forward wheel, rear wheel or both wheels on the verge of sliding.\\
\indent Time optimal paths obtained numerically were used to verify the 
path primitives during parallel parking, lane change and U-turn maneuvers. The examples support the paper finding that the front wheel $\cal {FW}$ path primitive usually dominates the rear wheel $\cal {RW}$ path primitive. The last example compared the time optimal {\small U}-turn paths under $\mu \!=\! 1.0$ and $\mu \!=\! 2.0$ coefficients of friction, showing that higher friction allows shorter travel time as one would expect, with the $\cal {RW}$ path primitive briefly appearing along the higher friction maneuver.\\
\indent Our current research seeks to deploy the time optimal path primitives of this paper to improve existing mobile robot motion planners. One important way would be to incorporate the path primitives into sampling based kinodynamic planners. These planners search the mobile robot state space along edges that represent dynamically feasible mobile robot paths. Current methods naively sample control actions, but the selection of control actions 
can now be informed by the path primitives and their 
approximate forward solutions described in this paper.
We also seek to incorporate the path primitives into machine learning planners: first generate data sets of time optimal maneuvers using direct numerical solvers, parse these maneuvers into their time~optimal path primitives, 
then train a~{\em path primitive classifier} that will predict the next path primitive and its switch time based on the mobile robot current state and current path primitive.

\indent Our longer term research will consider two extensions. The first extension would be to allow car-like robot navigation on variable slope terrains possibly with variable ground friction. Under this paper formulation, gravity will be multiplied by a~time varying slope parameter while the coefficient of friction will become piecewise constant. The second extension would be to consider car-like robot fuel or electricity consumption as an alternative path optimality criterion. Autonomous and semi-autonomous robotic cars are estimated to 
potentially save up to 
{\small 5\%-10\%} of fuel consumption when using energy optimal paths, thus offering hard-to-find fuel 
and pollution reduction opportunities.

\section{Appendix}

\noindent This appendix describes approximate analytic solutions for the $\cal {FW}$ and $\cal {RW}$ primitives and the solution for~the~$\cal {BW}$~primitive. The non-sliding constraints associated with $\cal {FW}$ and $\cal {RW}$ form convex regions  in the $(u_1,u_2)$ plane. Each of these regions
intersects an~individual control input limit 
at most along a~{\em single} segment for a~total of at most four segments.
The approximate solutions are taken as the endpoints of these segments as next described. 

Consider the contact force expressions from Section~IV: 
\[
\mbox{\small $
\begin{pmatrix}
F_{x_1} \\ F_{y_1}
\end{pmatrix}
= A_1(\nu,\phi) \! 
\begin{pmatrix} u_1 \\ u_2 \end{pmatrix}$}  + \mbox{\boldmath ${b}$}_1(\nu,\phi) %
\] 
\[
\mbox{\small $
\begin{pmatrix}  F_{x_2} \\ F_{y_2} \end{pmatrix}
= A_2(\nu,\phi) \! 
\begin{pmatrix} u_1 \\ u_2 \end{pmatrix} + \mbox{\boldmath ${b}$} _2(\nu,\phi) $} 
\]
\noindent where expressions for the $2 \!\times\! 2$ matrices
$A_1$ and~$A_2$ and the vectors $\mbox{\boldmath ${b}$}_1$ and $\mbox{\boldmath ${b}$}_2$
can be obtained from Eq.~\eqref{eq:forces}. 
First consider the approximate solutions for the 
$\cal {FW}$ primitive associated with $u_1 \!=\! \pm a_{max}$. Keeping $u_1$ fixed at its limits, an~active non-sliding constraint becomes the state dependent quadratic equation in the control input $u_2$:
\begin{equation} \label{eq.app}
\mbox{\hspace{-.65em}} \mbox{\small $ 
\begin{array}{lr}
\mbox{\small $\left( \!
A_1
\mbox{\small $\vctwo{\!\!\! \pm a_{max} \!\!\!}{\!\!\! u_2 \!\!\!}$}  \!+\!
\mbox{\boldmath ${b}$}_1 \!
\right)^T \!\!
\! \left( \!
A_1
\mbox{\small $\vctwo{\!\!\! \pm a_{max} \!\!\!}{\!\!\! u_2 \!\!\!} $} 
\!+\!
\mbox{\boldmath ${b}$}_1  \!
\right) $} & \\[10pt]
\mbox{\hspace{.2em}} \!=\! \mu^2 \big(\mbox{\small $\frac{1}{2}$}  g -  \tfrac{h}{d} \!\cdot\! ( \mbox{\boldmath $a$} \!\cdot\! 
\mbox{\small $\vctwo{\!\!\! \pm a_{max} \!\!\!}{\!\!\! u_2 \!\!\!}$}  \!+\! b) \big)^2  &
\end{array} 
$}
\end{equation}
where  $\mbox{\boldmath ${a}$}$ is the sum of the upper rows of  $A_1$~and~$A_2$~and~$b$ is the sum of the $x$-components of $\mbox{\boldmath ${b}$}_1$ and $\mbox{\boldmath ${b}$}_2$.
Now let $\Delta(\nu,\phi, \pm a_{max})$ be the discriminant of Eq.~\eqref{eq.app}. 
When $\Delta(\nu,\phi,  \pm a_{max}) \!\geq\! 0$, the two $u_2$ roots provide solutions $(-a_{max}, u^-_{2,I}(t))$, $(-a_{max}, u^-_{2,II}(t))$ for $u_1 \!=\! -a_{max}$ and $(a_{max}, u^+_{2,I}(t))$, $(a_{max}, u^+_{2,II}(t))$ for $u_1 \!=\! a_{max}$. Each of these solutions such that $|u_2| \leq\! b_{max}$ becomes one of the approximate time optimal solutions for the  $\cal {FW}$ primitive.

When $\Delta(\nu,\phi, -a_{max}) \!<\! 0$,  compute the {\em smallest} $u_1$-root of the discriminant equation $\Delta(\nu,\phi, u_1) \!=\! 0$. When $\Delta(\nu,\phi, a_{max}) \!<\! 0$, compute the {\em largest} $u_1$-root of the discriminant equation $\Delta(\nu,\phi, u_1) \!=\! 0$. In either case the respective root, $u^-_1$ or $u^+_1$, is the extreme $u_1$ value on the active non-sliding constraint. Now substitute back $u^-_1$ or $u^+_1$ into the quadratic non-sliding equations and solve for the single $u_2$-root of each equation, $u^-_2$ for $u^-_1$ or $u^+_2$ for $u^+_1$. The approximate time optimal solutions $(u^-_1(t),u^-_2(t))$ and  $(u^+_1(t),u^+_2(t))$ are taken when the corresponding 
$u_1 \!=\! - a_{max}$ or $u_1 \!=\! a_{max}$ limit lies outside the non-sliding constraint.

\begin{table}[t]
\centering
\renewcommand{\arraystretch}{1.3}
\begin{tabular}{|c|c|c|}
\hline
& $\mathcal{FW}$ & $\mathcal{RW}$ \\
\hline
$u_1 = -a_{max}$ &
\begin{tabular}{@{}c@{}}$(-a_{max}, u^-_{2,I}(t))$\\$(-a_{max}, u^-_{2,II}(t))$\end{tabular} &
\begin{tabular}{@{}c@{}}$(-a_{max}, u^-_{2,I}(t))$\\$(-a_{max}, u^-_{2,II}(t))$\end{tabular} \\
\hline
\begin{tabular}{@{}c@{}}$-a_{max}$ left \\ of $C_1$ or $C_2$\end{tabular} & $(u^-_1(t),u^-_2(t))$ & $(u^-_1(t),u^-_2(t))$ \\
\hline
$u_1 = +a_{max}$ &
\begin{tabular}{@{}c@{}}$(+a_{max}, u^+_{2,I}(t))$\\$(+a_{max}, u^+_{2,II}(t))$\end{tabular} &
\begin{tabular}{@{}c@{}}$(+a_{max}, u^+_{2,I}(t))$\\$(+a_{max}, u^+_{2,II}(t))$\end{tabular} \\
\hline
\begin{tabular}{@{}c@{}}$a_{max}$ right \\ of $C_1$ or $C_2$\end{tabular} & $(u^+_1(t),u^+_2(t))$ & $(u^+_1(t),u^+_2(t))$ \\
\hline
$u_2 = -b_{max}$ &
\begin{tabular}{@{}c@{}}$(u^-_{1,I}(t),-b_{max})$\\$(u^-_{1,II}(t),-b_{max})$\end{tabular} &
\begin{tabular}{@{}c@{}}$(u^-_{1,I}(t),-b_{max})$\\$(u^-_{1,II}(t),-b_{max})$\end{tabular} \\
\hline
\begin{tabular}{@{}c@{}}$-b_{max}$ below \\ $C_1$ or $C_2$\end{tabular} & $(u^-_1(t),u^-_2(t))$ & $(u^-_1(t),u^-_2(t))$ \\
\hline
$u_2 = +b_{max}$ &
\begin{tabular}{@{}c@{}}$(u^+_{1,I}(t),+b_{max})$\\$(u^+_{1,II}(t),+b_{max})$\end{tabular} &
\begin{tabular}{@{}c@{}}$(u^+_{1,I}(t),+b_{max})$\\$(u^+_{1,II}(t),+b_{max})$\end{tabular} \\
\hline
\begin{tabular}{@{}c@{}}$b_{max}$ above \\ $C_1$ or $C_2$\end{tabular} & $(u^+_1(t),u^+_2(t))$ & $(u^+_1(t),u^+_2(t))$ \\
\hline
\end{tabular}
\caption{Approximate solutions for $\mathcal{FW}$ and $\mathcal{RW}$ path primitives valid when $|u^{\pm}_{1,I}(t)|,|u^{\pm}_{1,II}(t)| \leq a_{max}$ and $|u^{\pm}_{2,I}(t)|,|u^{\pm}_{2,II}(t)| \leq b_{max}$.}
\end{table}


A similar procedure computes approximate solutions for the $\cal {FW}$ primitive associated with $u_2 \!=\! \pm b_{max}$.  Keeping $u_2$ fixed at its limits,
consider the quadratic equations in the control input $u_1$:
\begin{equation}  \label{eq.app1}
\mbox{\hspace{-.65em}} \mbox{\small $ 
\begin{array}{lr}
\mbox{\small $\left( \!
A_2
\vctwo{\!\!\! u_1 \!\!\!}{\!\!\! \pm b_{max} \!\!\!} \!+\!
\mbox{\boldmath ${b}$}_2 \!
\right)^T \!\!
\! \left( \!
A_2
\vctwo{\!\!\! u_1 \!\!\!}{\!\!\! \pm b_{max} \!\!\!} 
\!+\!
\mbox{\boldmath ${b}$}_2  \!
\right) $} & \\[10pt]
\mbox{\hspace{.2em}} \!=\! \mu^2 \big(\mbox{\small $\frac{1}{2}$}  g +  \tfrac{h}{d} \!\cdot\! ( \mbox{\boldmath $a$} \!\cdot\! \vctwo{\!\!\! u_1 \!\!\!}{\!\!\! \pm b_{max} \!\!\!} \!+\! b) \big)^2  &
\end{array} 
$}
\end{equation}
where $\mbox{\boldmath ${a}$}$ and $b$ are the same as above. Solving Eq.~\eqref{eq.app1} gives the solutions $(u^-_{1,I}(t),-b_{max})$, $(u^-_{1,II}(t),-b_{max})$ for $u_2 \!=\! -b_{max}$ and $(u^+_{1,I}(t),b_{max})$, $(u^+_{1,II}(t),b_{max})$~for~\mbox{$u_2 \!=\! b_{max}$,} or possibly $(u^-_1(t),u^-_2(t))$ and  $(u^+_1(t),u^+_2(t))$ when the corresponding $u_2$ limit lies outside the non-sliding constraint. Table~III summarizes 
the approximate solutions for the $\cal{FW}$ primitive as well as
the approximate solutions for the $\cal{RW}$ primitive computed in a similar manner. 

\begin{figure}
\centering
\includegraphics[width=0.65\linewidth]{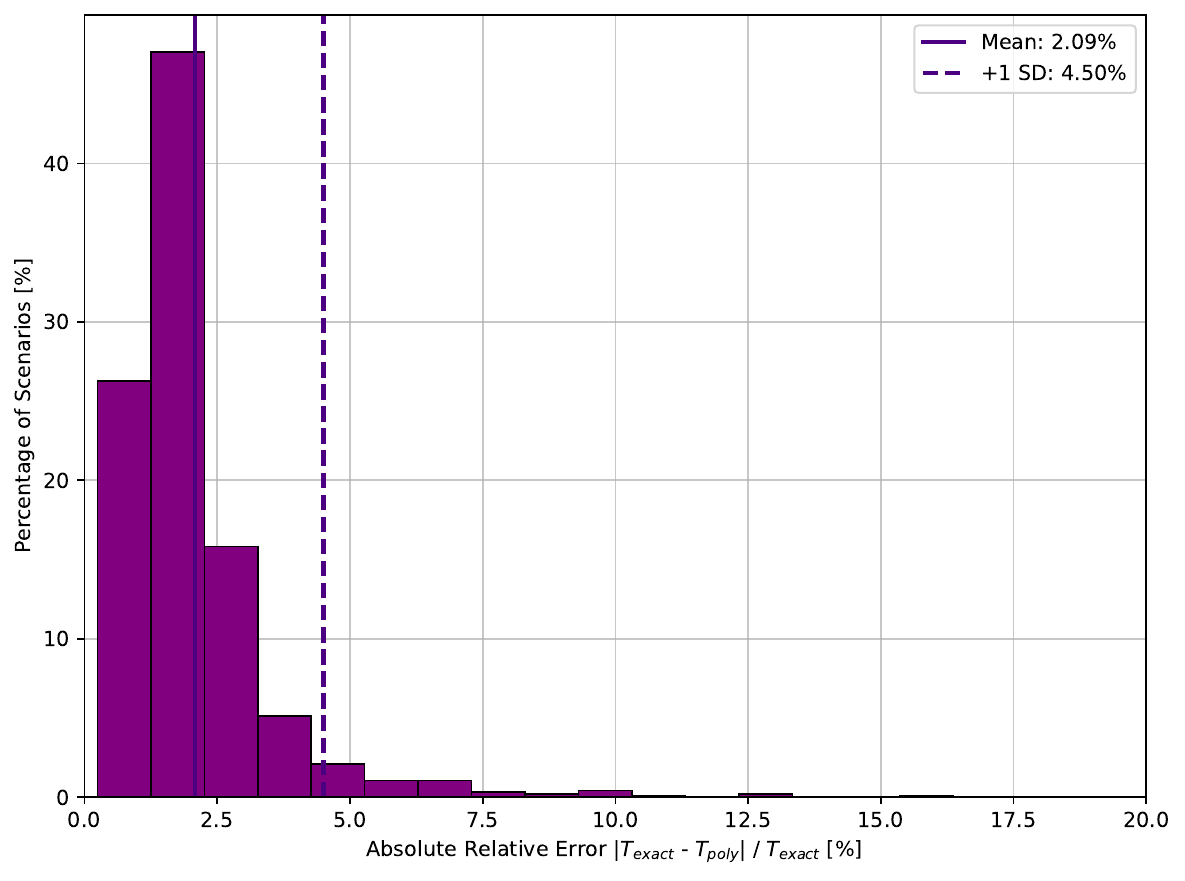}
\caption{Histogram of travel time along approximate time optimal paths computed over $1,000$ random start and target states.  Mean travel time is $2.09\%$ longer than the optimal travel time  with standard deviation of $4.5\%$.}
	\label{fig:hist}
\end{figure}

Table~III gives up to eight approximate solutions that can be interpreted as sample points on the respective non-sliding constraint. In practice, we found that retaining a fixed number of eight sample points  gives much tighter approximation of the optimal travel time. Details of this practical extension as pseudo-code and the code itself can be found in Ref.~\cite{leeor_code}. The resulting approximation quality is shown in Fig.~\ref{fig:hist} as the histogram of approximate travel time computed over $1,000$ random start and target states of the car-like robot. The mean travel time along the approximate paths is roughly $2\%$ longer than the optimal travel time,  with standard deviation of $4.5\%$.

\begin{figure}
\centering
\includegraphics[width=0.65\linewidth]{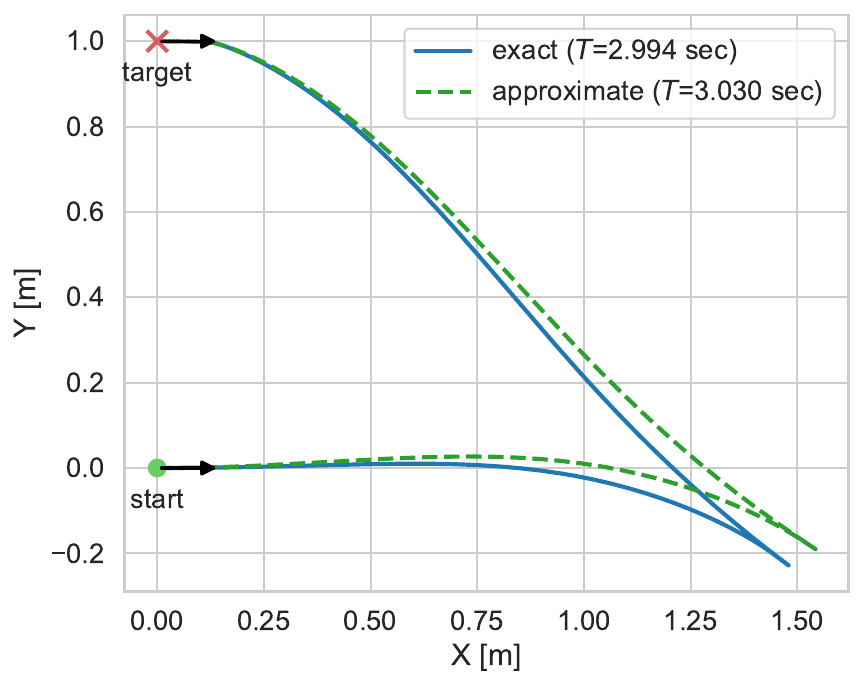} 
\caption{Approximate time optimal path (dashed curve) and exact time optimal path  (solid curve) for the parallel parking maneuver of Example~1.}
	\label{fig:App}
\end{figure}

{\bf Example:} Fig.~\ref{fig:App} shows
the exact and approximate time optimal paths  for 
the parallel parking maneuver of Example~1. The  approximate path travel time of $3.03$~seconds is slightly longer than the travel time of $2.994$~seconds along the true time optimal path. The example uses the approximate solutions of Table~III augmented into eight approximate solutions as described above. Pseudo-code and the code running this example appear in Ref.~\cite{leeor_code}.~\eex

Finally, consider the exact solution for the $\cal{BW}$ primitive where both wheels are on the verge of sliding. One can verify that the rear wheel non-sliding constraint forms an~ellipse for center-of-mass heights $h \!\leq\! 2L$. Under this condition, one can  
parametrize the rear wheel non-sliding ellipse, then substitute the parametrization into the front wheel  non-sliding constraint which is quadratic, then solve the resulting quartic equation for the four possible solutions of the $\cal{BW}$ primitive. One can alternatively add a~homogeneous coordinate and obtain the 
$\cal{BW}$ solutions as roots of two quadratic equations.


\bibliographystyle{IEEEtran}
\bibliography{elonbib,elon,books}

\end{document}